\documentclass[10pt,twocolumn,letterpaper]{article}

\usepackage[pagenumbers]{cvpr}

\usepackage[final]{microtype}

\usepackage{xspace}
\newcommand{\METHOD}{LAMP\xspace}
\newcommand{\lamp}{LAMP\xspace}

\usepackage{booktabs}
\usepackage{pifont}
\usepackage{siunitx}
\usepackage{tikz}
\usetikzlibrary{arrows.meta}
\usepackage[accsupp]{axessibility}

\definecolor{cvprblue}{rgb}{0.21,0.49,0.74}
\usepackage[pagebackref,breaklinks,colorlinks,allcolors=cvprblue]{hyperref}

\def\paperID{****}
\def\confName{CVPR}
\def\confYear{2026}

\title{LAMP: Localization Aware Multi-camera People Tracking in Metric 3D World}

\author{Nan Yang\hspace{.12in}
Julian Straub\hspace{.12in}
Fan Zhang\hspace{.12in}
Richard Newcombe\hspace{.12in}
Jakob Engel\hspace{.12in}
Lingni Ma\\
\vspace{0.05in}\\
Meta Reality Labs Research
}

\begin{document}
\twocolumn[{
  \renewcommand\twocolumn[1][]{#1}%
  \maketitle
  \begin{center}
    \captionsetup{type=figure}
    \begin{tikzpicture}[inner sep=0]
        \node(p00) {\includegraphics[height=60mm]{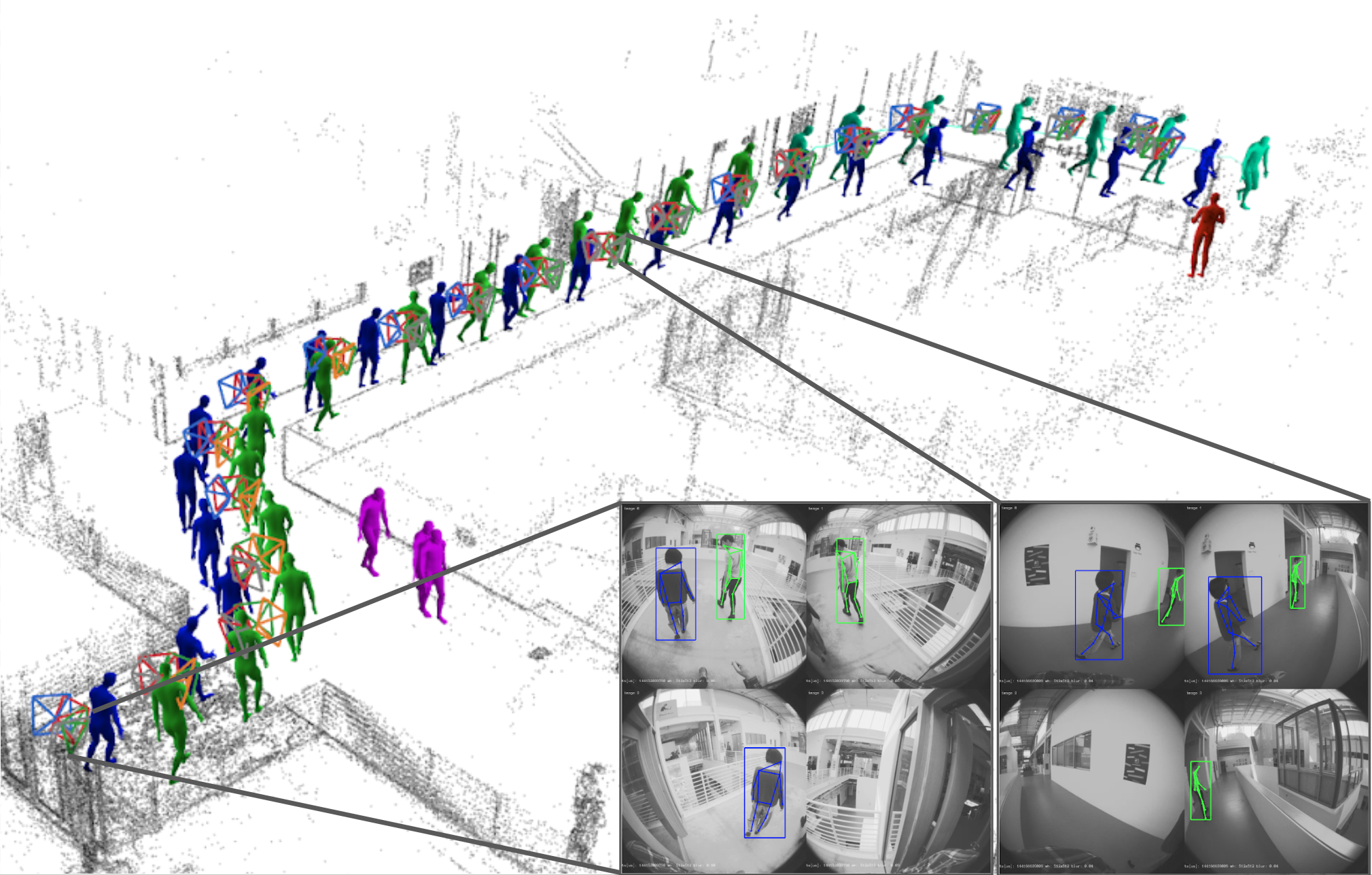}};
        \node (p10) at(p00.north east)[xshift=6px, anchor=north west]{\includegraphics[height=29mm]{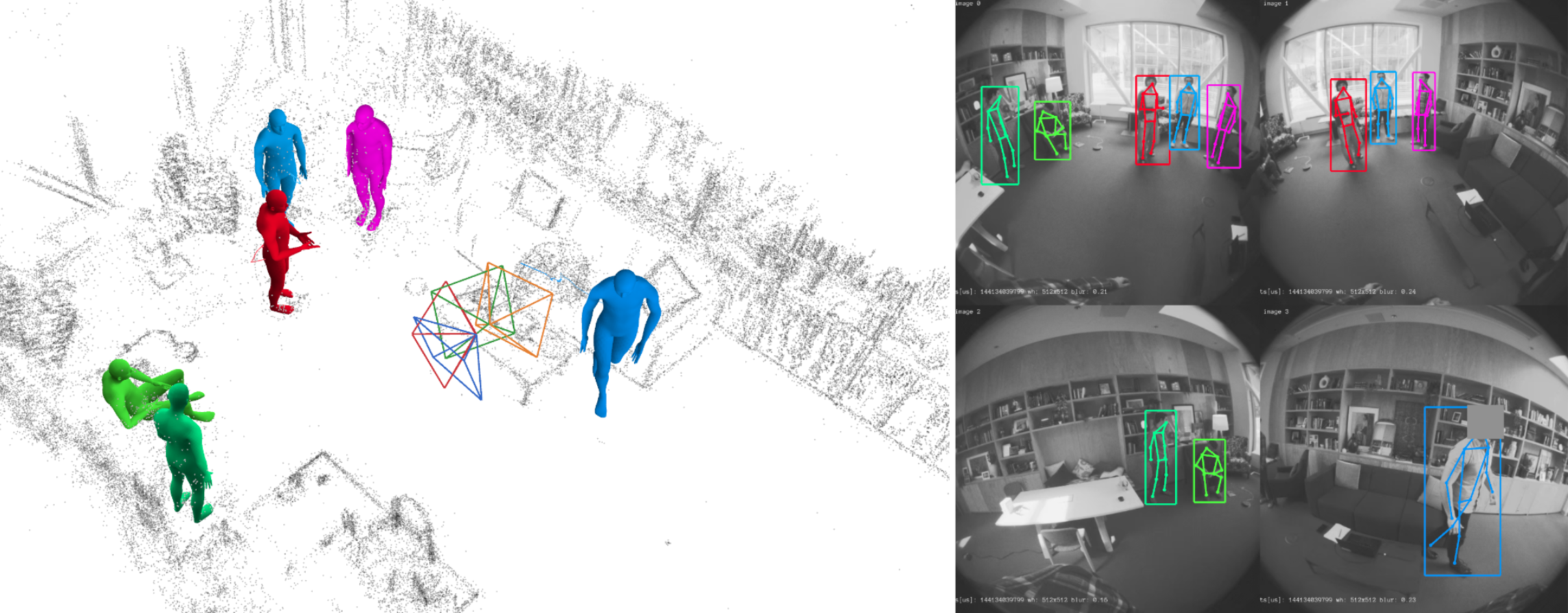}};
        \node (p11) at(p00.south east)[anchor=south west, xshift=6px]{\includegraphics[height=29mm]{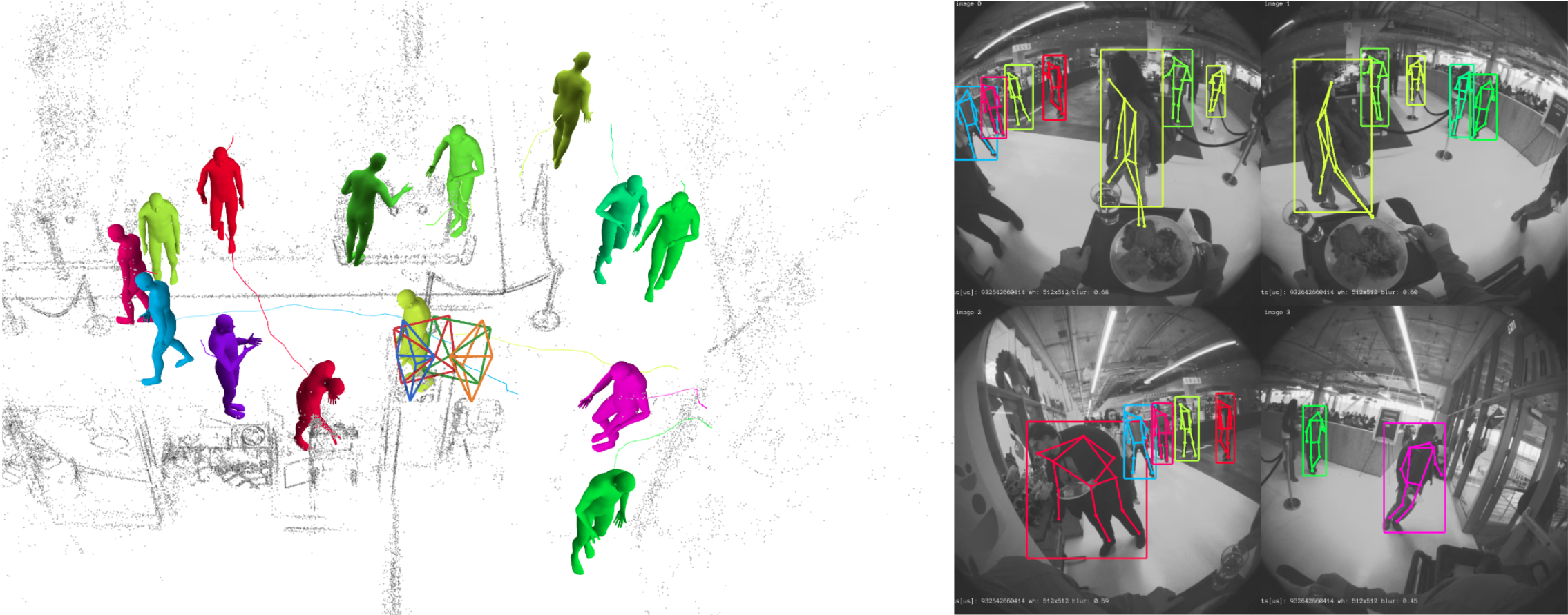}};
        \draw [draw=gray, inner sep=0pt](p00.south west) rectangle (p00.north east);
        \draw [draw=gray, inner sep=0pt](p10.south west) rectangle (p10.north east);
        \draw [draw=gray, inner sep=0pt](p11.south west) rectangle (p11.north east);
    \end{tikzpicture}

    \captionof{figure}{\textbf{We propose \lamp, the first method to track human motion with multi-camera headsets in the metric 3D world}.
        \textit{(Left)} \lamp persistently track the body motion over a long time, where the 2D observations constantly switching between different cameras across time.
        \textit{(Right)} \lamp tracks multiple people simultaneously in real-time across a ultra-wide field-of-view by combining all cameras. Please refer to the \emph{supplementary video} for the real-time demonstrations.}
    \label{fig:teaser}
  \end{center}
}]

\begin{abstract}
 Tracking 3D human motion from egocentric multi-camera headset is challenged by severe egomotion, partial visibility or occlusions and lack of training data. 
 Existing methods designed for monocular video often require static or slowly-moving cameras and cannot efficiently leverage multi-view, calibrated and localized input. 
 This makes them brittle and prone to fail on dynamic egocentric captures. 
 We propose LAMP (\textbf{L}ocalization \textbf{A}ware \textbf{M}ulti-camera \textbf{P}eople Tracking): a novel, simple framework to solve this via early disentanglement of observer and target motion. 
 LAMP introduces a two-step process. First, we leverage the known device 6~DoF motion and calibration to convert detected 2D body keypoints from all cameras over a temporal window into a unified 3D world reference frame. 
 Second, an end-to-end-trained spatio-temporal transformer fits 3D human motion directly to this 3D ray cloud. This "lift-then-fit" approach allows \lamp to learn and leverage a natural human motion prior in the world-space, 
 as well as providing an elegant framework to flexibly incorporate information from multiple temporally asynchronous, partially observing and moving cameras. 
 LAMP achieves state-of-the-art results on monocular benchmarks, while significantly outperforming baselines for our targeted egocentric setting. Project page: \url{https://facebookresearch.github.io/LAMP}.
\end{abstract}

\vspace{-6mm}

\section{Introduction}
\label{sec:intro}

Augmented reality and smart glasses~\cite{AppleVisionPro,engel2023project,AriaGen2} promise to be persistent, context-aware assistants, enhancing daily life by deeply understanding the user's environment and acting as effective front-ends that connect a wearer to powerful AI systems. A critical component of such contextual AI systems is social understanding, i.e., the ability to perceive people around the user, what they are doing, and how they are interacting with the user or each other. Understanding 3D human motion is a fundamental component to understanding these higher-level social cues.

Tracking 3D human body motion from videos has been intensively studied over the past decades with impressive progress~\cite{hmmr,hmr2,wham,shen2024gvhmr,wang2025prompthmr,wang2024tram}.
However, tracking people observed by egocentric modern headsets~\cite{MetaQuest3, AppleVisionPro, HoloLens2, engel2023project} presents novel challenges that often render existing methods ineffective, due to three major factors.
First, headsets are subject to \textit{significant 6-DoF egomotion} from constant, rapid head movement of wearers.
This breaks many state-of-the-art motion tracking algorithms, which assume or heavily rely on static or slow-moving camera motion~\cite{bewley2016simple,zhang2022bytetrack,newell2025comotion}.
More fundamentally, these methods attempt to track the motion between observer and target -- which can work well when the observer intentionally follows or captures the target (correlated motion), but breaks when these are uncorrelated and subject to independent motion patterns.
Second, modern egocentric headsets~\cite{MetaQuest3, AppleVisionPro, HoloLens2, engel2023project, AriaGen2} are designed to be \textit{multiview camera rigs}.
In order to cover a large field of view, each individual camera captures a different viewing direction, with partial stereo overlapping to a subset of all cameras, as shown in Fig.~\ref{fig:mv_track_ex}.
This means, a person may be fully or partially observed by a single or multiple cameras at each point in time,
while the observations also switch from one camera to another across the field-of-view over time.
These properties are poorly considered by many existing solutions, which are designed for monocular input.
Consequentially, partial single-view observations lead to frequent tracking loss,
camera hand-off scenarios are ineffective handled via late fusion,
and the results are subject to scale ambiguity inherit to all monocular 3D tracking problems.
Third, most existing methods demand a large-scale video data annotated with 3D human motion for training.
This type of data is sparse and requires tremendous efforts to collect~\cite{ego_body, khirodkar2023egohumans,khirodkar2024harmony4d},
while synthetic data~\cite{bedlam,yin2024whac} often lacks realistic moving camera motions. At the same time, modern headsets update camera configurations in every new generations and across manufacturers -- making it impractical to built sufficiently large training datasets for a specific device. This is the core reason why most methods that rely on raw video pixels as input to ML models focus on monocular, un-posed and un-calibrated input as lowest common denominator.

To address these challenges, we introduce \lamp: an \textit{early world-space ray lifting} paradigm to track multiple people in metric 3D world with modern egocentric headsets.
\lamp takes a posed multi-view video clip as input, and outputs the 3D parameterized body pose~\cite{smpl} per timestamp per person.
To achieve this, the method first detects the 2D keypoints~\cite{vitpose,wu2019detectron2} independently per person per camera per timestamp.
The 2D keypoints are then back-projected and posed by the 6-DoF camera poses into 3D rays, and associated by the targets' identity across time.
The resulting grouped 3D ray cloud is processed by \lamp-Net, a spatial-temporal transformer to output the parameterized body motion for each timestamp.
By using the posed 3D rays as input to train our model, \lamp achieves two explicit factorization.
First the headset motion is factored out from tracking targets' motion by using the 6-DoF device localization and camera calibrations,
which are both available for modern AR or VR headsets by running highly optimized the state-of-the-art visual-inertial odometry (VIO) or full SLAM systems~\cite{engel2023project,engel2018dso,mourikis2007msckf,orb2} achieving remarkable accuracy and reliability~\cite{krishnan2025lamaria}.
This factorization enables the entire subsequent fitting problem to fully focus on learning the prior of how people move,
and naturally stabilizes and anchors the estimated body motion in the metric 3D world.
Second, 2D keypoints detection on raw image pixels is factored out clearnly from 3D motion fitting, thus allowing to leverage existing 2D, image-based human and keypoints detections algorithms~\cite{detr, coco, lin2014microsoft}.
This factorization naturally handles partial observations by individual cameras, and allows for seamless cross-camera hand-off scenarios to derive consistent tracking result. Further, it allows to re-use datasets across different rig-layouts:
Since \lamp-Net only requires 2D keypoint observations as input,
we can simulate the training data for arbitrary multi-camera headset configurations from arbitrary existing motion datasets~\cite{amass, ma2024nymeria}.
The contribution of our work is summarized as follows.

\begin{itemize}[leftmargin=*]
\item We introduce \lamp, a novel system that \emph{lifts and tracks multiple people} directly in the metric 3D world frame from egocentric multi-camera videos by exploiting known 6-DoF camera poses from modern headsets.
\item We propose an \emph{early world-space ray lifting} formulation that factors out headset egomotion by lifting 2D keypoints to 3D ray cloud before spatio-temporal reasoning, enabling training from simulated data and naturally supporting multi-view inputs.
\item We demonstrate that \lamp significantly outperforms monocular approaches, effectively leveraging multi-view and posed inputs to improve both tracking accuracy and field-of-view coverage. We further show that \lamp \emph{matches} state-of-the-art monocular baselines when reduced to a monocular input rig for a fair comparison.
\item We intend to publicly release our model and code to facilitate future research.
\end{itemize}

\section{Related Work}
\label{sec:related_work}

\begin{figure*}
    \centering
   \includegraphics[width=0.98\linewidth]{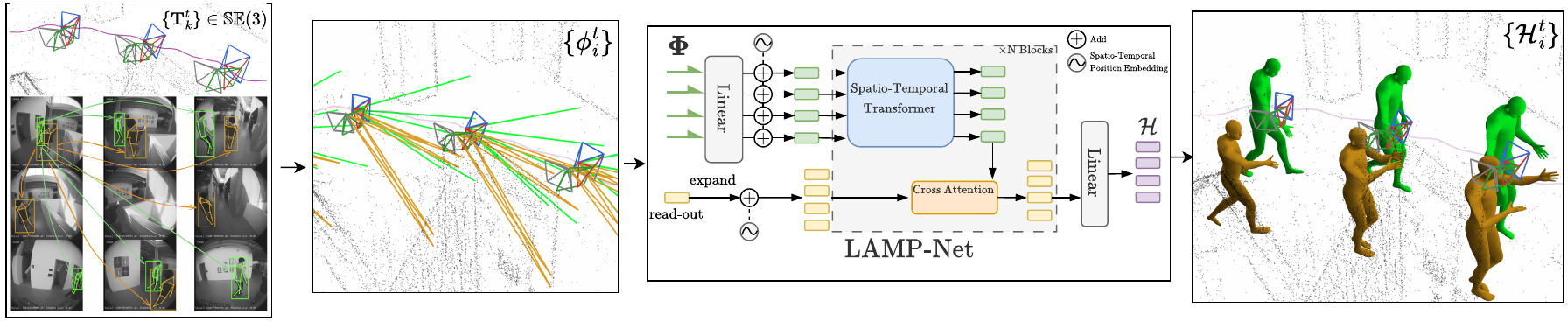}
    \caption{\textbf{Method overview.} From egocentric multi-camera video with known 6-DoF poses $\{\mathbf{T}_{k}^{t}\!\in\!\mathrm{SE}(3)\}$, we detect 2D boxes and keypoints
and associate them over time to form per-person tracks as shown in the first subplot.
Using the known intrinsics/extrinsics, the 2D keypoints of each track are \emph{lifted}
to a sequence of spatio-temporal posed \textit{3D ray clouds} in a gravity-aligned reference frame as shown in the second subplot.
We then train the \lamp-Net to perform spatio-temporal reasoning which maps ray cloud $\{\boldsymbol{\phi}_i^t\}$ to world-grounded body motion $\{\mathcal{H}_{i}^{t}\}$ as shown in the last subplot.}
  \label{fig:system_overview}
\end{figure*}

\paragraph{Single image 3D human pose estimation.} A large body of work focuses on predicting 3D human pose from a single 2D image~\cite{wang2014robust, pavlakos2017coarse, bogo2016keep, moreno20173d, pavlakos2018learning, martinez2017simple}. Early works~\cite{opt_smplify, sminchisescu2002human} employ a classical optimization approach using a parametric human model, such as SMPL~\cite{smpl} and SMPL-X~\cite{smplx} and minimize the energy with respect to the image observations, such as 2D keypoints and silhouettes. With the advance of deep learning, regression approaches~\cite{hmr,hmr2,cliff,patel2025camerahmr,kolotouros2019learning} become more popular as they learn human pose and shape priors from large datasets~\cite{bedlam,amass,h36m}, delivering strong generalization capabilities and robustness. However, these single-image models focus on estimating the 3D human pose in the camera coordinate frame, without considering the camera movement or temporal consistencies.

\paragraph{Video-based 3D human pose estimation.} Temporal cues are widely exploited to improve stability and reduce per-frame ambiguity in 3D human motion estimates~\cite{transformer,pavllo20193d,zhang2022mixste,mehta2017vnect,arnab2019exploiting,zhou2016sparseness,li2022mhformer,zheng20213d,vibe}. Most pipelines extract \emph{per-frame} features and then aggregate them over time using temporal convolutions~\cite{hmmr,pavllo20193d,zhang2022mixste}, recurrent models~\cite{vibe,meva}, or Transformers~\cite{wan2021encoder,shen2023global,li2022mhformer,zhu2023motionbert}. While effective for enforcing temporal smoothness, the majority of these methods reason in the \emph{camera} coordinate system and do not explicitly account for camera motion or maintain a consistent world reference, making them ill-suited for scenarios with substantial 6-DoF egomotion (e.g., head-mounted cameras).

\paragraph{Multi-view 3D pose estimation.} Many methods reconstruct 3D human pose from multiple, synchronized, and calibrated static cameras, typically via triangulation or volumetric feature fusion \cite{zhang2021direct, kocabas2019self, srivastav2024selfpose3d, remelli2020lightweight, wu2021graph, lin2021multi, chen2020multi, opt_multiview, qian2025bevtrack, vo2020self}. These approaches assume a fixed multi-camera rig set up, and sufficient spatial coverage to maintain visibility, , which limits their applicability beyond studio-style setups. In contrast, our scenario uses a \emph{moving} multi-camera rig captured from head mounted devices that casually captures the environment with rapid camera motions.

\paragraph{World-coordinate 3D human pose estimation.}
Estimating human motion in a fixed world frame is essential for perceiving and interacting with the physical environment. Recent work has begun addressing world-frame recovery from moving cameras. Optimization-based methods such as SLAHMR~\cite{slahmr} and PACE~\cite{pace} refine poses and trajectories but are ill-suited for real-time use. Similarly, MVLift~\cite{li2025lifting} proposes a diffusion network to synthesize multi-view observations from single view 2D skeleton videos, followed by optimization to fit 3D skeletons, which is not real-time capable. WHAM~\cite{wham} proposes an end-to-end neural model that leverages gyroscope signals to predict world-frame poses. GVHMR~\cite{shen2024gvhmr} introduces a gravity-view coordinate to stabilize orientation and reduce drift by predicting in a gravity-aligned frame. To obtain metric scale, TRAM~\cite{wang2024tram}, WHAC~\cite{yin2024whac}, and PromptHMR~\cite{wang2025prompthmr} couple off-the-shelf SLAM~\cite{droid,teed2024dpvo} with monocular depth~\cite{zoedepth}, then transform camera-frame predictions into the world frame. GloPro~\cite{schaefer2023glopro} assumes known camera poses but still predicts in camera space before transforming; Ray3D~\cite{zhan2022ray3d} employs a ray-based representation under static-camera rigs.  In contrast, we \emph{explicitly} exploit the readily available 6-DoF camera poses on egocentric devices~\cite{AppleVisionPro,AriaGen2,engel2023project} and adopt an \textbf{early-lifting} paradigm that lifts the 2D keypoints to 3D rays expressed in a local world frame \emph{before} spatio-temporal reasoning. This avoids late composition as the prior works, enables causal real-time inference, and lets the model learn a unified 3D human motion prior in the world coordinate system.

\section{Method}\label{sec:method}

Given multi-view videos $\{I_k^t\}_K^T$ captured by a multi-camera headset with $k\in[1,\ldots, K]$ cameras over $t\in[1,\ldots, T]$ timestamps,
\lamp outputs the metric 3D grounded \textbf{body motion tracklets} for each person observed by the video grounded in 3D world coordinates.
We assume the camera calibration and 6~DoF camera poses $\mathbf{T}_{k}^t \in\mathbb{SE}(3)$ are known and accurate.
To parameterize 3D human motion, we adopt the commonly used SMPL~\cite{smpl} format.
For each tracked person $i$ at time $t$, the body pose is denoted by
$\mathcal{H}_i^t:=\{\boldsymbol{\theta}_i^t, \boldsymbol{\beta}_i^t, \boldsymbol{\omega}_i^t, \boldsymbol{\tau}_i^t\}$,
where $\boldsymbol{\theta}_i^t\in\mathbb{R}^{69}$ is the SMPL pose parameters that represents the joint angles of $23$ body joints,
$\boldsymbol{\beta}_i^t\in\mathbb{R}^{10}$ is the SMPL shape parameters with the first 10 PCA coefficients,
$\boldsymbol{\omega}_i^t\in\mathbb{SO}(3)$ is the global 3D rotation of the root joint at pelvis,
$\boldsymbol{\tau}_i\in\mathbb{R}^3$ is the global 3D translation of the root joint.
We further denote the clip of human motion by $\mathcal{H}_i^{T}:= \{\mathcal{H}_i^{t}\}_{t=1}^{T}$
With the function $f_\text{SMPL}(\boldsymbol{\theta}, \boldsymbol{\beta})$ that converts SMPL shape and pose parameters to $\mathcal{J}\in\mathbb{R}^{23}$ joints and $\mathcal{V}\in\mathbb{R}^{6890}$ mesh vertices in SMPL local coordinates,
the joints and vertices in the world coordinates for person $i$ at time $t$ are then computed by
\begin{equation}\label{eq:smpl}
    \mathcal{J}_i^t, \mathcal{V}_i^t = \text{exp}(\boldsymbol{\omega}_i^t) f_\text{SMPL}(\boldsymbol{\theta}_i^t, \boldsymbol{\beta}_i^t) +\boldsymbol{\tau}_i^t \;.
\end{equation}

\subsection{Overview}
\Cref{fig:system_overview} illustrate how \lamp employs an \textit{early world-space ray lifting} paradigm to track multiple people over time for the multi-camera headset.
Start from individual images, the method first detects 2D bounding boxes of each observed person and estimate the set of 2D keypoints $\{\mathbf{p}_j^t\in\mathbb{R}^2\}$ per bounding box detection.
The 2D detections are spatio-temporally associated across $K$ cameras over $T$ timestamps by the identity of tracking targets.
After associations, the 2D keypoints are back-projected into 3D and transformed by the known camera poses $\mathbf{T}_k^t$ and camera calibration parameters to obtain a sequence of spatio-temporal ray clouds aligned in the 3D space.
For each group of ray clouds, we use \lamp-Net to map the sequence of 3D rays into 3D grounded human motion $\{\mathcal{H}_i^t\}$.

In this formulation, we made two explicit factorizations.
First, the 6~DoF camera poses are disentangled from the movements of tracked people.
We argue this is a crucial ingredient which enables \lamp-Net to fully focus on learning the motion prior of how people move.
Unlike many previous works that are designed to learn or refine a coarse camera motion together with human motion~\cite{wang2025prompthmr,wang2024tram},
we believe learning camera motion not only make the problem harder to solve, but also is unnecessary for modern headsets, which device tracking is largely considered a solved problem~\cite{engel2023project,krishnan2025lamaria,davison2007monoslam,orb,engel14ECCV,mourikis2007msckf,leutenegger2022okvis2,forster2014svo}. The second factorization is to separate 2D human detections from the motion learning and directly learn to solve the inverse problem ``find the parameterized 3D human motion that is most consistent with a given spatio-temporal ray cloud''.
The formulation encourages \lamp-Net to flexibily combine and leverage temporally asynchronous observations for ``3D triangulation'' by connecting them through the local dynamics imposed by human motion prior.
By doing so, \lamp is able to naturally handle partial 2D observations and camera hand-off scenarios to persist tracking across time and camera views.
As a by-product of this factorization, \lamp can simply be re-trained for a new camera configuration by simulating 2D observations for the targeted camera layout from arbitrary motion datasets. This solves a key challenge in data scarcity for posed multi-view devices, which is the core reason why most research focuses on the monocular setting.
In the remaining section, we further describe the ray association, 2D-to-3D multi-view lifting and \lamp-Net training for spatio-temporal ray fitting in more details.

\begin{figure}
  \centering
  \includegraphics[width=0.99\columnwidth]{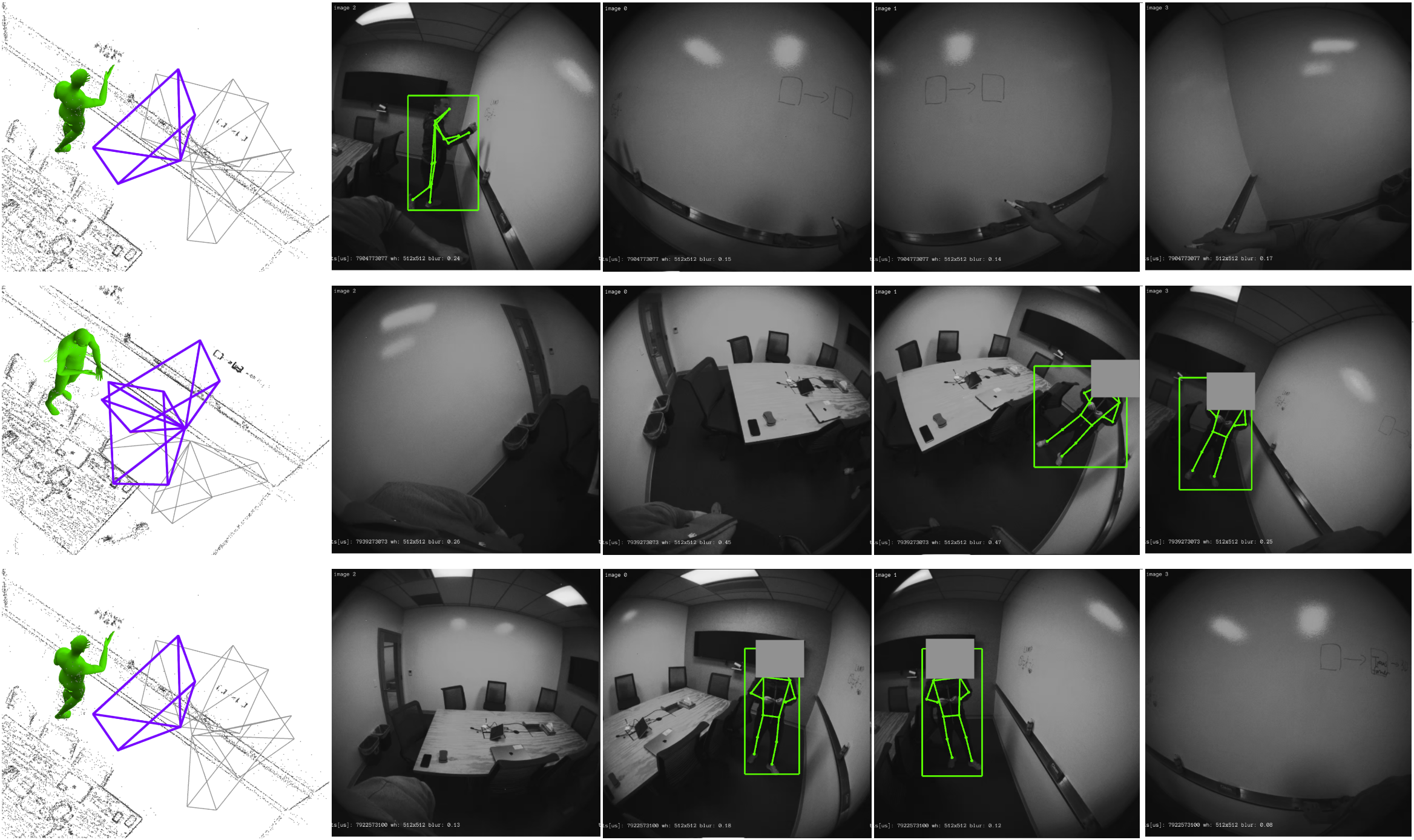}
  \caption{\textbf{Multi camera tracking. } \lamp~seamlessly estimates body motion for a person across several ``camera-handoffs'' for a sequence captured from an Project Aria Gen2 glasses and using all four available monochrome cameras. Our formulation allows to seamlessly combine all available observations in a single model inference call to fitting a full $4s$ 3D motion snippet.}
    \label{fig:mv_track_ex}
\end{figure}

\subsection{Tracklet spatio-temporal association}
For a timestep $t$, we obtained the 2D bounding boxes of all visible people using 2D detection algorithms~\cite{wu2019detectron2,redmon2016you,yolox2021}.
For association, we solve a \textit{bipartite matching problem} between these 2D bounding boxes and active tracklets from the previous timesteps.
This is done by projecting 3D points into the images to obtain an expected observation location and compute the matching cost.
Note that this automatically compensates for the headset motion, as tracklets are represented in the world coordinates.
Matching observations is then solved with the Hungarian algorithms~\cite{Kuhn1955,zhang2022bytetrack,bewley2016simple}.
When a 2D bounding box fails to associate with others, we instantiate a new tracklet.
For real world inference, a tracklet will be deactivated if the person is not observed for more than a threshold of time.

\subsection{World aligned ray lifting}
For each 2D bounding box, we run ViTPose~\cite{vitpose} to detect the MSCOCO~\cite{lin2014microsoft} 2D keypoints on the cropped images.
We then assemble the 3D ray cloud ${\boldsymbol{\phi}_i^t}$ by first unprojecting the 2D keypoints into 3D,
transforming them by the corresponding camera pose,
and normalizing the reference coordinates by the first frame of the temporal window to prepare the input for \lamp-Net inference.
For coordinates normalization,
we define a local coordinates $L_t$ based on the pose of the camera $i=0$, which effectively aligns the reference with gravity as estimated by visual inertial odometry (VIO)
\begin{equation}
\mathbf{T}^t_{W\leftarrow L_T} := \text{gravity\_align}(\mathbf{T}^t_{W\leftarrow C_0}) \in \mathbb{SE}(3).
\end{equation}
Together, the compute of 3D rays into the local coordinate system follows
\begin{equation}
    ^{c}\boldsymbol{\phi}_j^t := \mathbf{T}_{L_T \leftarrow W} \cdot \mathbf{T}^t_{W \leftarrow C_k}  \cdot \pi^{-1}(\mathbf{p}_j^t),
\end{equation}
where $\pi^{-1}\colon\Omega\to\mathbb{R}^3$ denotes the calibrated camera un-projection function of a pixel to a unit ray, and $\mathbf{T}^t_{W \leftarrow C_k}\in \mathbb{SE}(3)$ denotes the known camera pose of camera $k$ at time $t$. Combined with the respectively transformed camera centers
\begin{equation}
    ^{o}\boldsymbol{\phi}_t^k := \mathbf{T}_{L_T \leftarrow W} \cdot \mathbf{T}^t_{W \leftarrow C_k}  \cdot [0,0,0,1]^\top,
\end{equation}
this results in one 3D ray per 2D observation. Finally, we stack all rays within the window, parametrized as 6-dimensional Pl\"ucker rays concatenated with the confidence score from the 2D keypoint detector.
The final result is a tensor $\boldsymbol{\Phi} \in \mathbb{R}^{{T \times K}\times J \times 7}$, where
$J=17$ is the number of MSCOCO human keypoints and K is the number of cameras. The values corresponds to the missing 2D keypoints are set to zero for training and inference with \lamp-Net, described in the next section.

\subsection{Spatio-temporal ray fitting with \lamp-Net}
The function to map spatio-temporal 3D ray cloud to the corresponding 3D human motion is modeled by \lamp-Net.
To this end, we use a spatio-temporal transformer architecture~\cite{zhu2023motionbert,vaswani2017attention}. The transformer takes the sequential ray cloud as the input, and regresses the SMPL motion parameters~\cite{smpl} $\{\mathcal{H}_i^t\}$ for each timestamp.
The spatio-temporal encoder performs self-attention across both spatial (i.e., by joint) and temporal (i.e., by frame) dimensions to jointly estimate body shape and motion dynamics.
A learnable read-out embedding is expanded and added with the time encoding to form the query tokens in a cross-attention decoder.
Unlike previous works where network readouts interact only with the last encoder layer~\cite{goel2023humans}, our decoder performs cross-attention at each encoder block, allowing the readout embeddings to iteratively aggregate motion and geometry information across different feature hierarchies.
We found that this multi-level interaction improves temporal stability and convergence.
For training, we adopt the 6D rotation~\cite{Zhou_2019_CVPR} to parameterize 3D rotation. Note that since we normalize the input by the local coordinates $L_T$, the predicted motions needs to be transformed back to the original 3D world reference.

\paragraph{Training losses.}
Inspired by prior work~\cite{wang2024tram, hmr2}, the network training losses penalize the deviation to the ground-truth SMPL parameters $\mathcal{H}_i^t$, the 3D joint positions $\mathcal{J}$, 3D vertex positions $\mathcal{V}$, as well as the joint velocity $\mathcal{D}_J$. The loss function is defined as
\begin{equation}\textstyle
\mathcal{L} = \lambda_\text{SMPL}\mathcal{L}_\text{SMPL} + \lambda_\text{3D}\mathcal{L}_\text{3D} +  \lambda_\text{V}\mathcal{L}_\text{V} + \lambda_\text{vel}\mathcal{L}_\text{vel} \;,
\label{eq:loss}
\end{equation}
with
\begin{align*}
\textstyle\mathcal{L}_\text{SMPL} &= \textstyle\frac{1}{T}\sum_{t=1}^{T}||\hat{\mathcal{H}}^t - \mathcal{H}^t||^2_2 \\
\mathcal{L}_\text{3D} &=\textstyle \frac{1}{T}\sum_{t=1}^{T}||\mathcal{\hat{J}}_{t} - \mathcal{J}_{t}||^2_F \\
\mathcal{L}_\text{V} &= \textstyle\frac{1}{T}\sum_{t=1}^{T}||\hat{\mathcal{V}}_{t} - \mathcal{V}_{t}||^2_F \\
\mathcal{L}_\text{vel} &= \textstyle\frac{1}{T-1}\sum_{t=2}^{T}||\hat{\mathcal{D}}_{t} - \mathcal{D}_{t}||^2_F \;.
\label{eq:terms}
\end{align*}
We empirically found that the vertices loss $\mathcal{L}_\text{V}$ improves the results even though it is ill-posed for \METHOD to recover accurate mesh due to the sparse 2D keypoint observations.

\begin{figure}[t]
  \centering
   \includegraphics[width=0.98\linewidth]{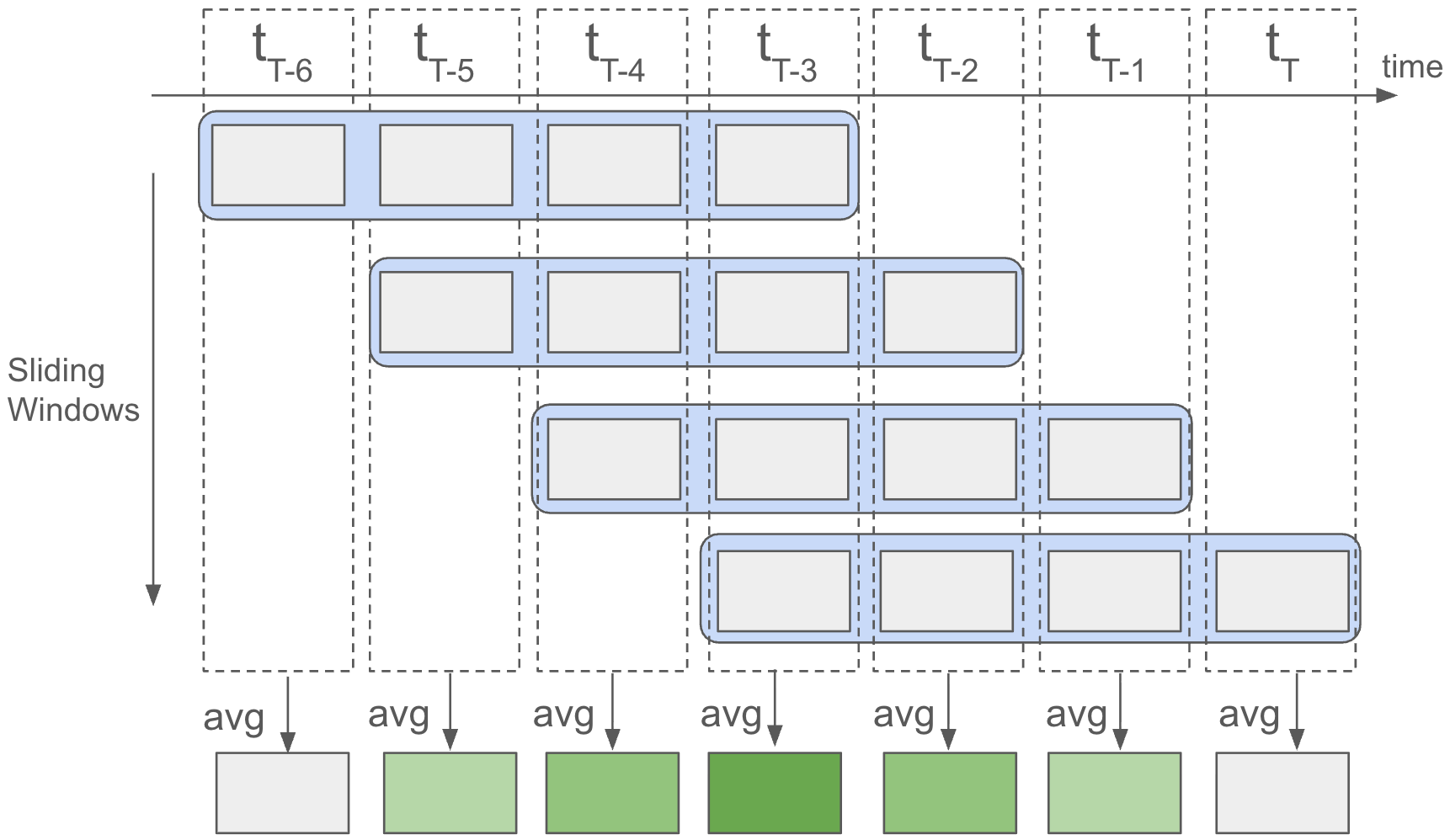}
   \caption{\textbf{Sliding window inference}. We propose a simple and light-weight sliding-window inference strategy by averaging the same-time-stamp pose predictions to get more accurate and stable human motions.}
   \label{fig:sliding_window}
\end{figure}

\subsection{Non-causal inference with temporal smoothing}
\lamp-Net consumes a window of video clips as input, each frame will be processed $T$ times during a strictly causal online inference, where the input window increments forward 1 frame at a time to yield the estimate for the latest frame.
This inference paradigm is illustrated in \cref{fig:sliding_window}.
With this nature of inference, smoothing the past estimations by averaging the predictions over time comes with no additional cost network inference per se.
Assume the network predictions contain noise, averaging multiple estimates over time is expected to improve the accuracy and reduce jitter.
However, smoothing results over sliding window increases the latency of output. At maximum, it can delay the output by $T-1$ timestamps.
In practice, this property of \lamp inference provides a runtime variable that can be used control the tradeoff between accuracy versus latency.
In our experiments, we evaluate the impact of smoothing with maximum latency, where the pose at time $t$ is computed by averaging all $T$ variants of predictions available over time.

\section{Experiments}
\label{sec:experiments}

\begin{figure*}
    \centering
    \def\xs{2pt}
    \begin{tikzpicture} [inner sep=0pt,font=\footnotesize]
    \node(p0) {\includegraphics[width=0.49\textwidth]{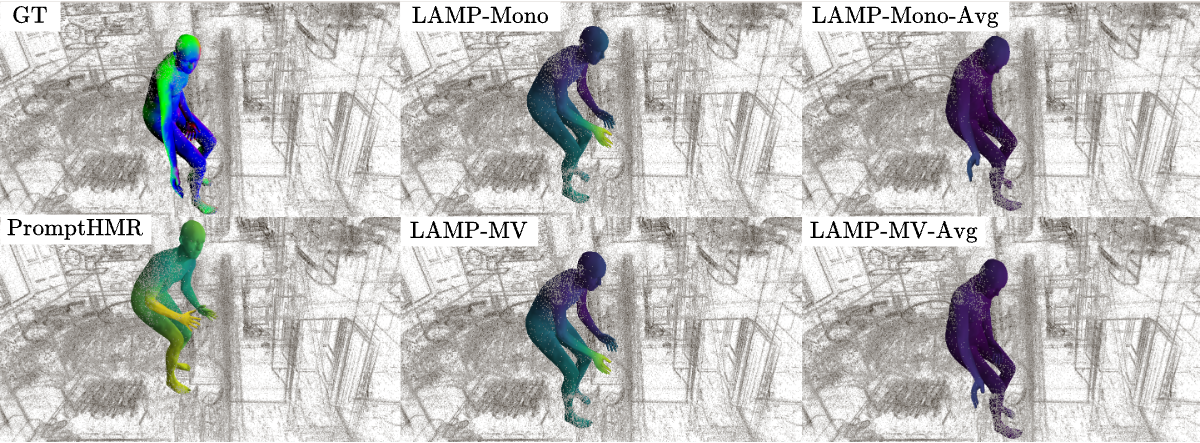}};
    \node(p1) at(p0.east)[anchor=west,xshift=\xs] {\includegraphics[width=0.49\textwidth]{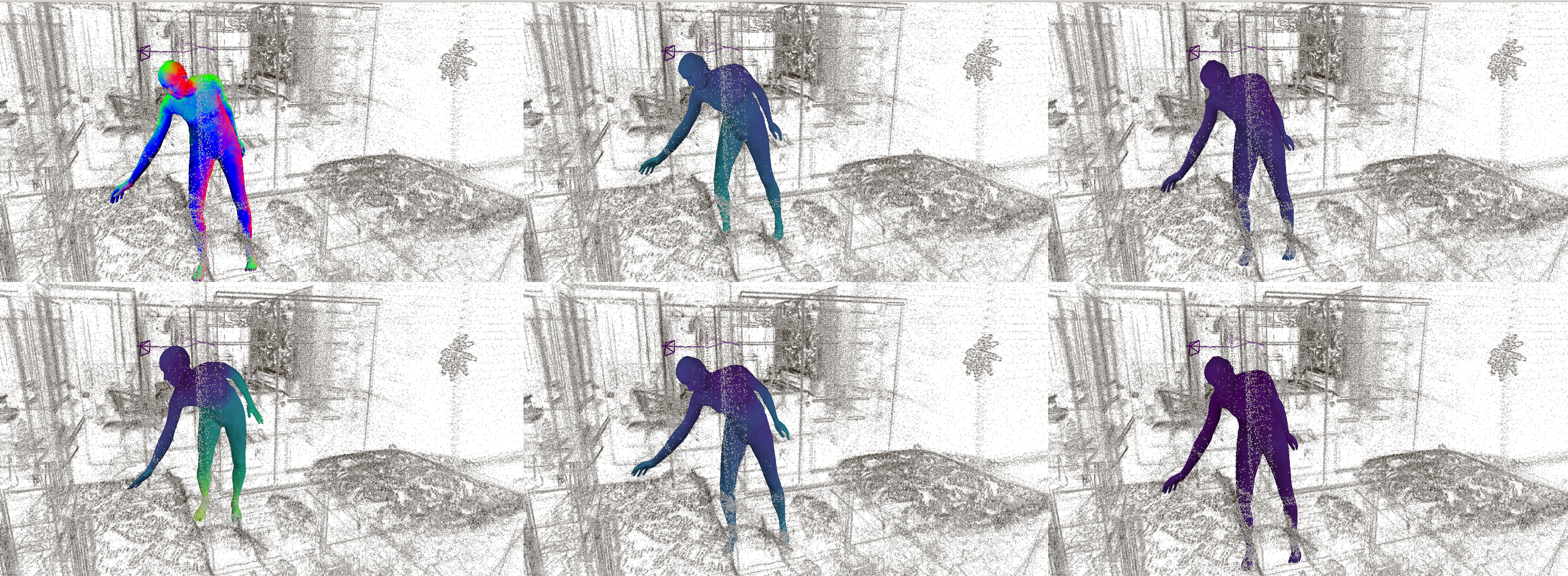}};
    \node(p2) at(p0.south)[anchor=north, yshift=-2pt] {\includegraphics[width=0.49\textwidth]{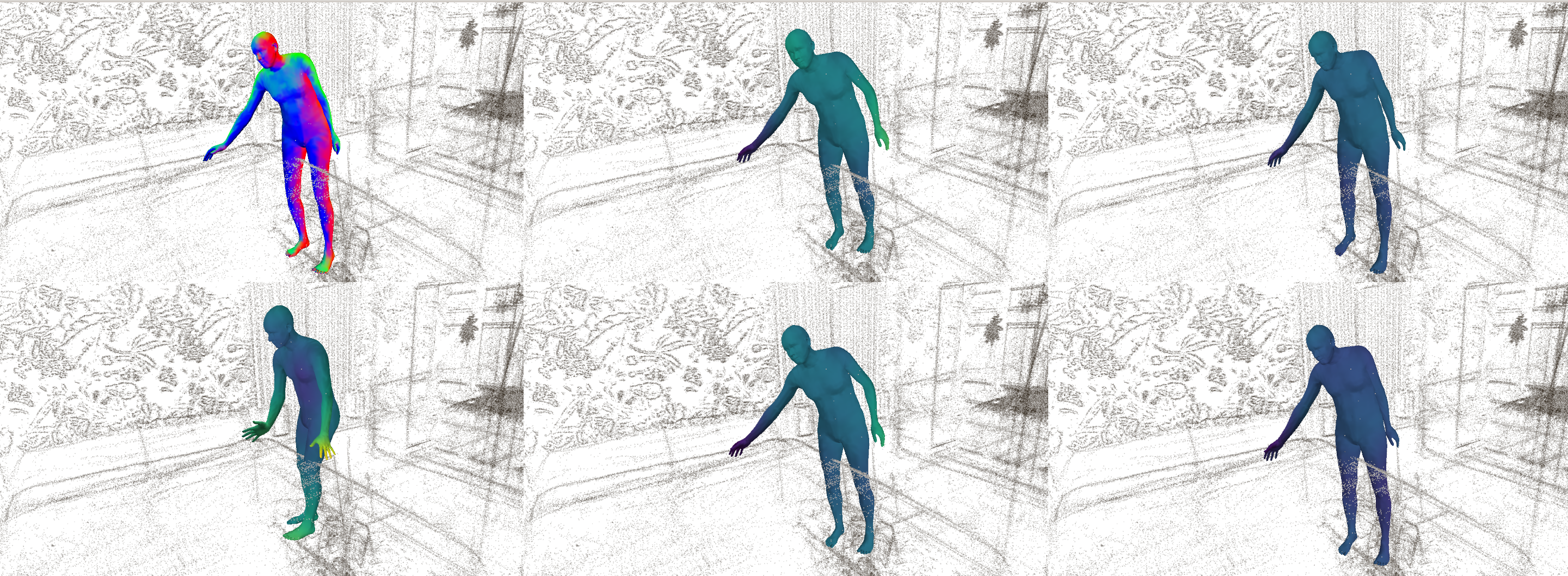}};
    \node(p3) at(p1.south)[anchor=north,yshift=-2pt] {\includegraphics[width=0.49\textwidth]{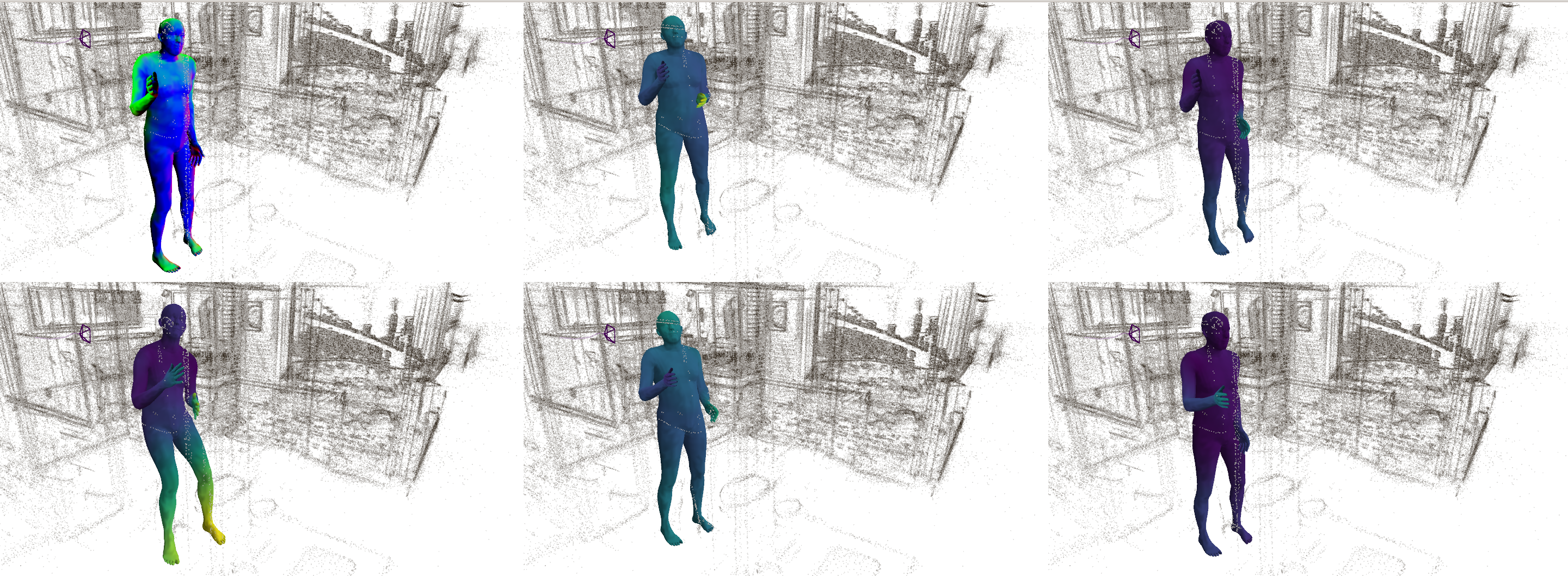}};
    \draw [inner sep=0pt](p0.south west) rectangle (p0.north east);
    \draw [inner sep=0pt](p1.south west) rectangle (p1.north east);
    \draw [inner sep=0pt](p2.south west) rectangle (p2.north east);
    \draw [inner sep=0pt](p3.south west) rectangle (p3.north east);

    \end{tikzpicture}
    \vskip-3pt
    \caption{\textbf{Qualitative comparison of 3D human motion estimation.}
    We compare the output PromptHMR~\cite{wang2025prompthmr} against monocular \lamp and multi-camera (MV) \lamp on Nymeria~\cite{ma2024nymeria} with and without temporal smoothing.
    Per output SMPL mesh, the vertex colors encode the Euclidean distance to the corresponding ground-truth vertex, with the higher error showed in yellow and lower error in dark purple.}
   \label{fig:visual_cmp}
\end{figure*}
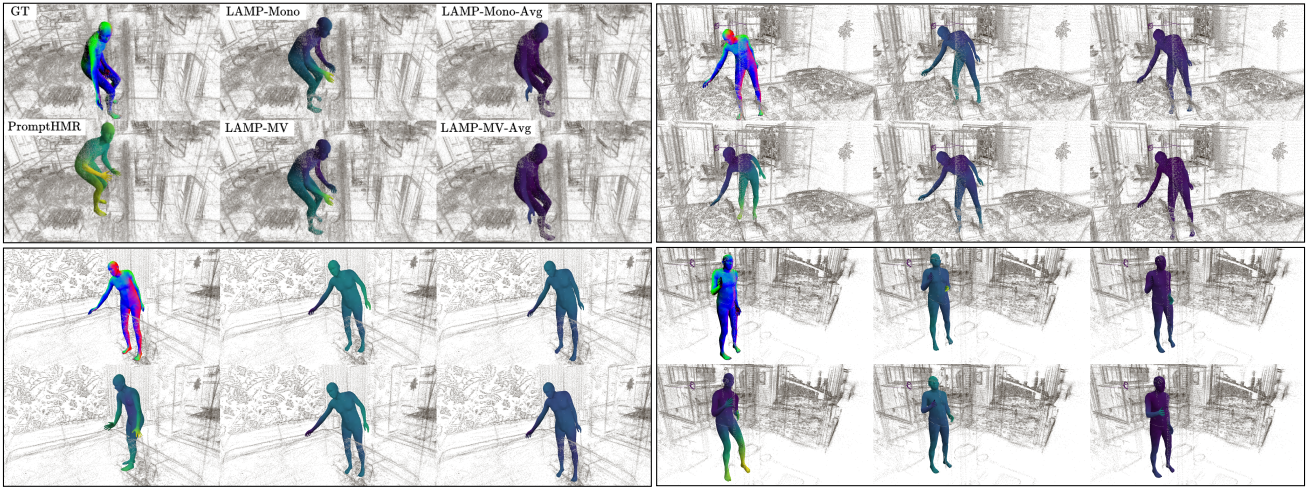

\begin{figure*}
\centering
\def\ih{23mm}
\begin{tikzpicture}[inner sep=0pt,font=\footnotesize]
    \node(p0) {\includegraphics[height=\ih]{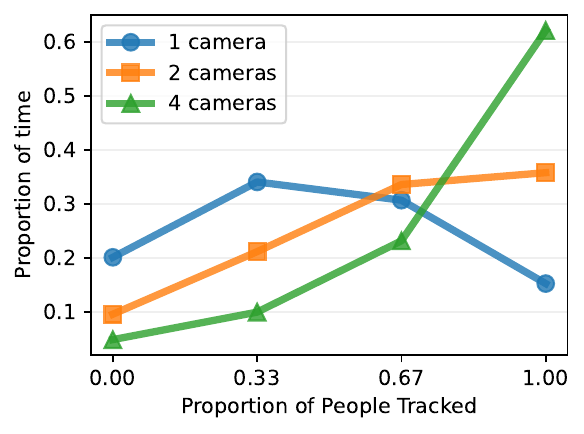}};
    \node(p1) at(p0.east)[anchor=west,xshift=1pt] {\includegraphics[height=\ih]{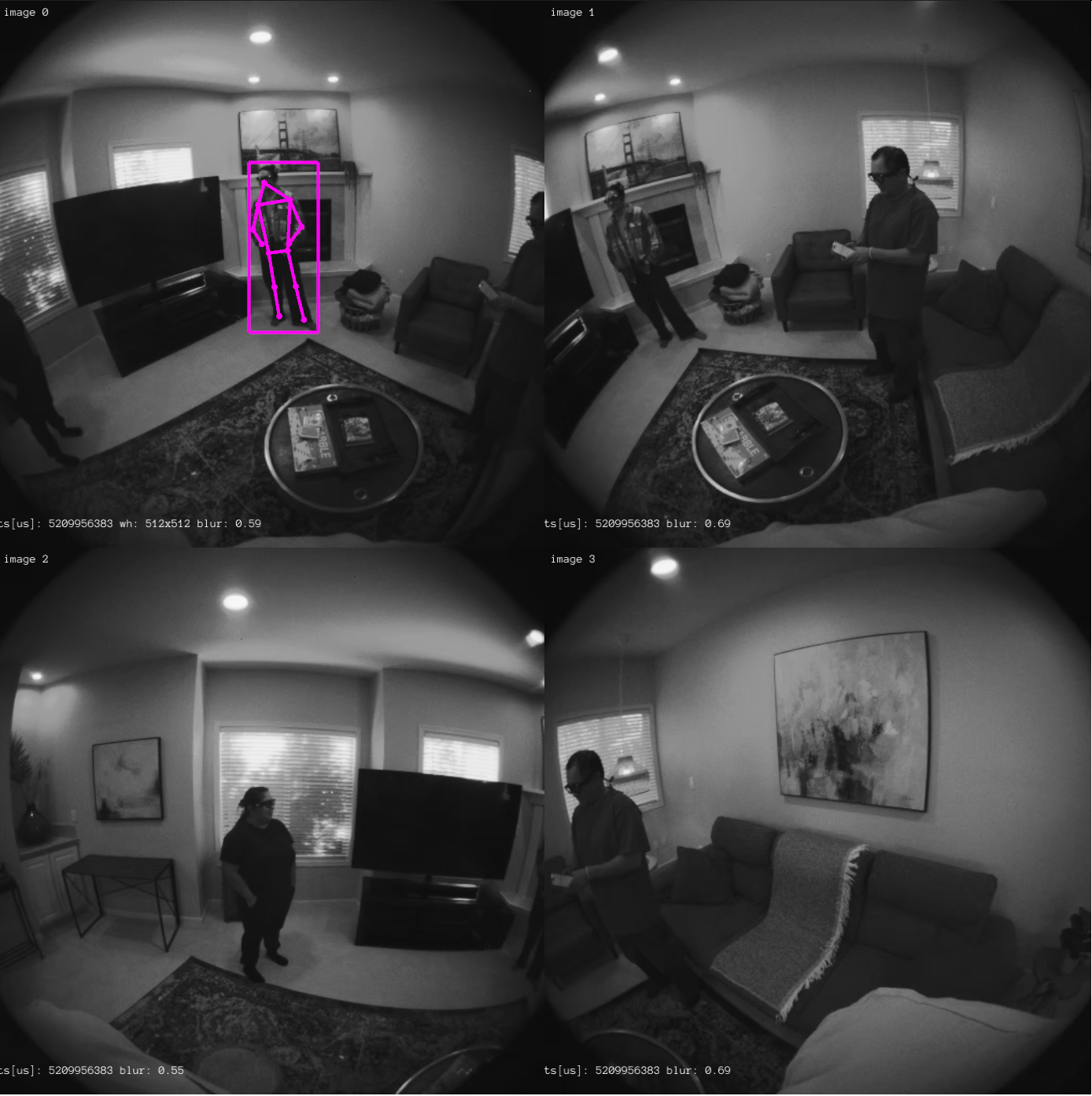}};
    \node(p2) at(p1.east)[anchor=west,] {\includegraphics[height=\ih]{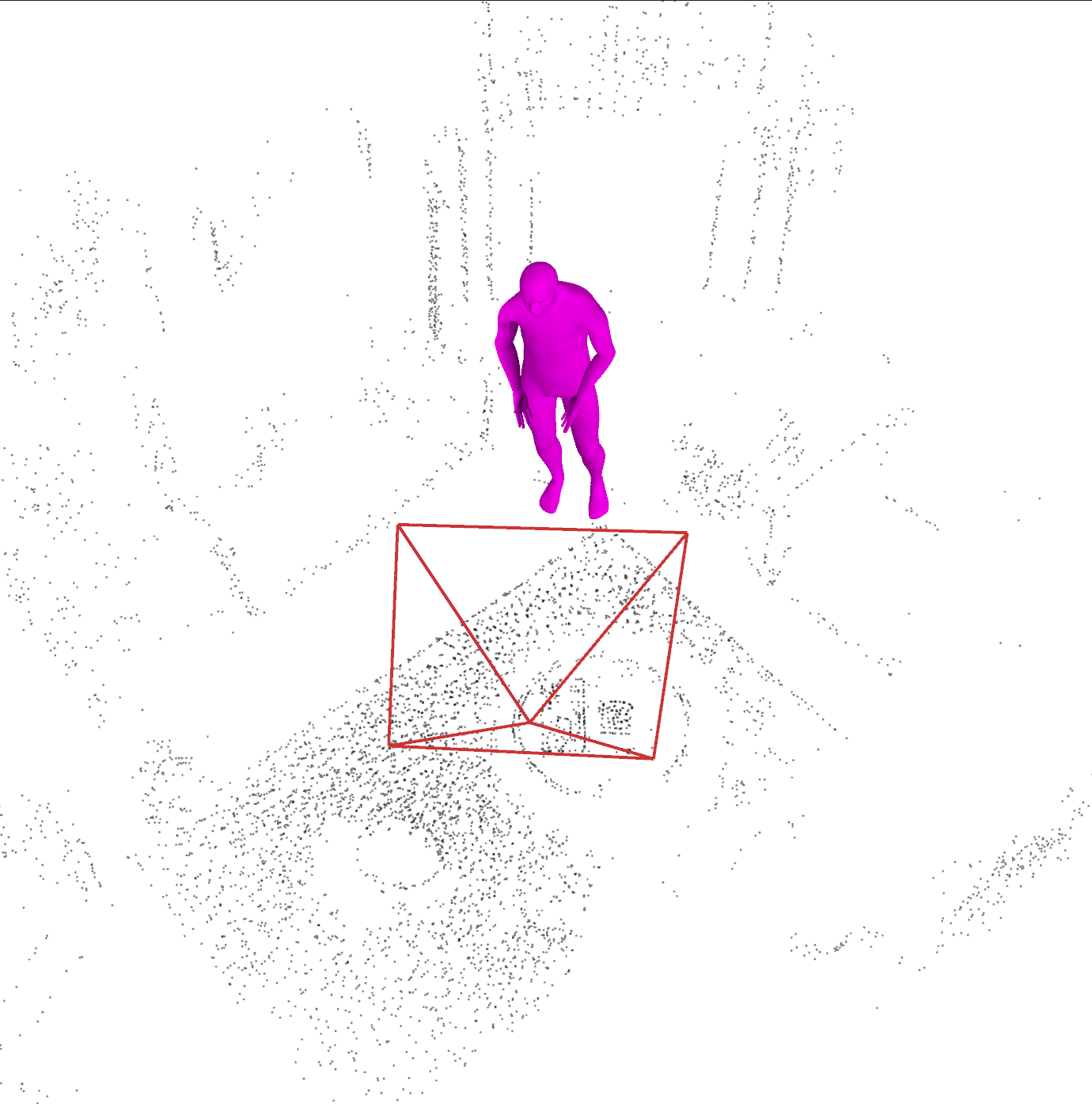}};
    \node(p3) at(p2.east)[anchor=west, xshift=1pt] {\includegraphics[height=\ih]{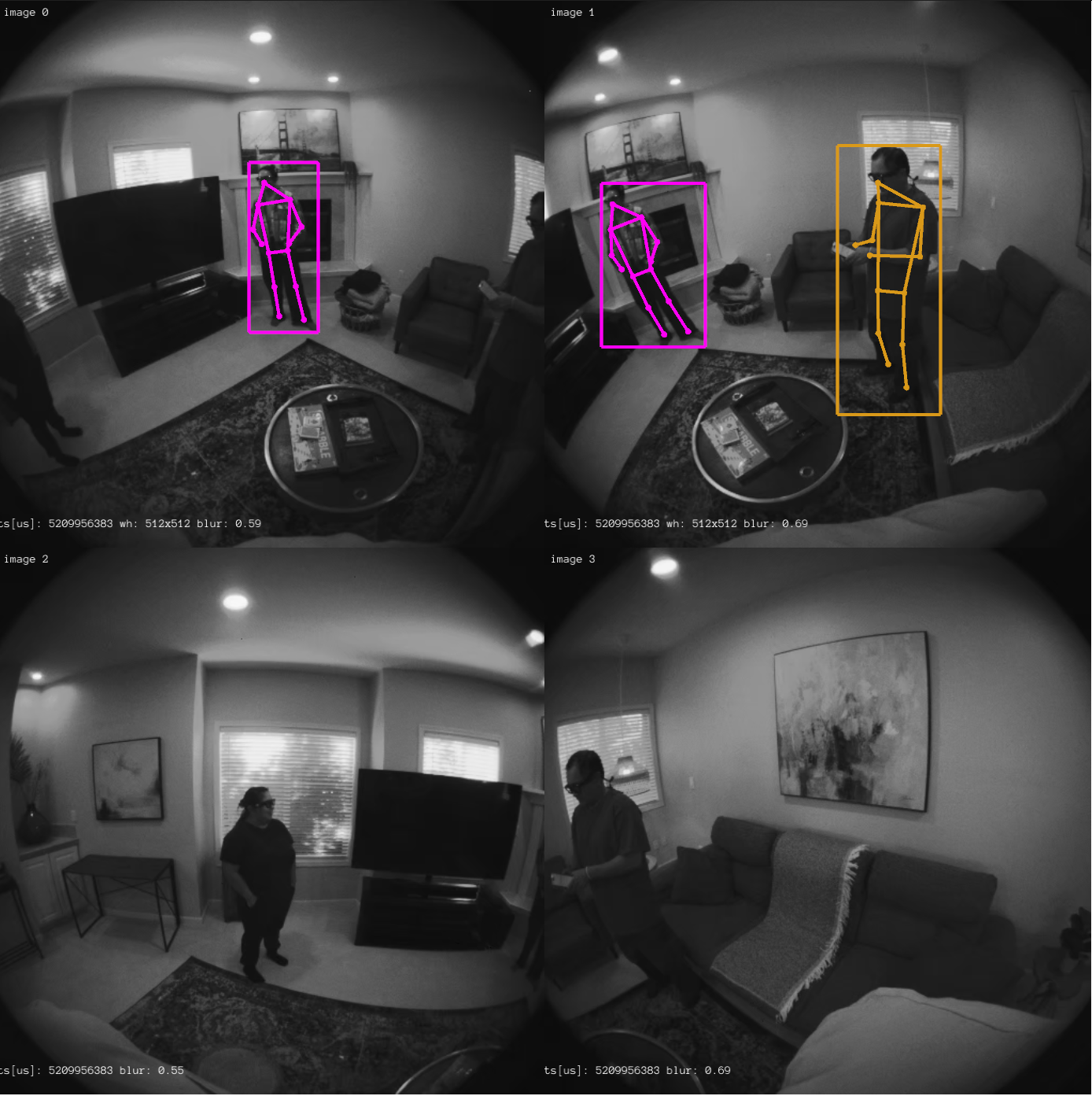}};
    \node(p4) at(p3.east)[anchor=west] {\includegraphics[height=\ih]{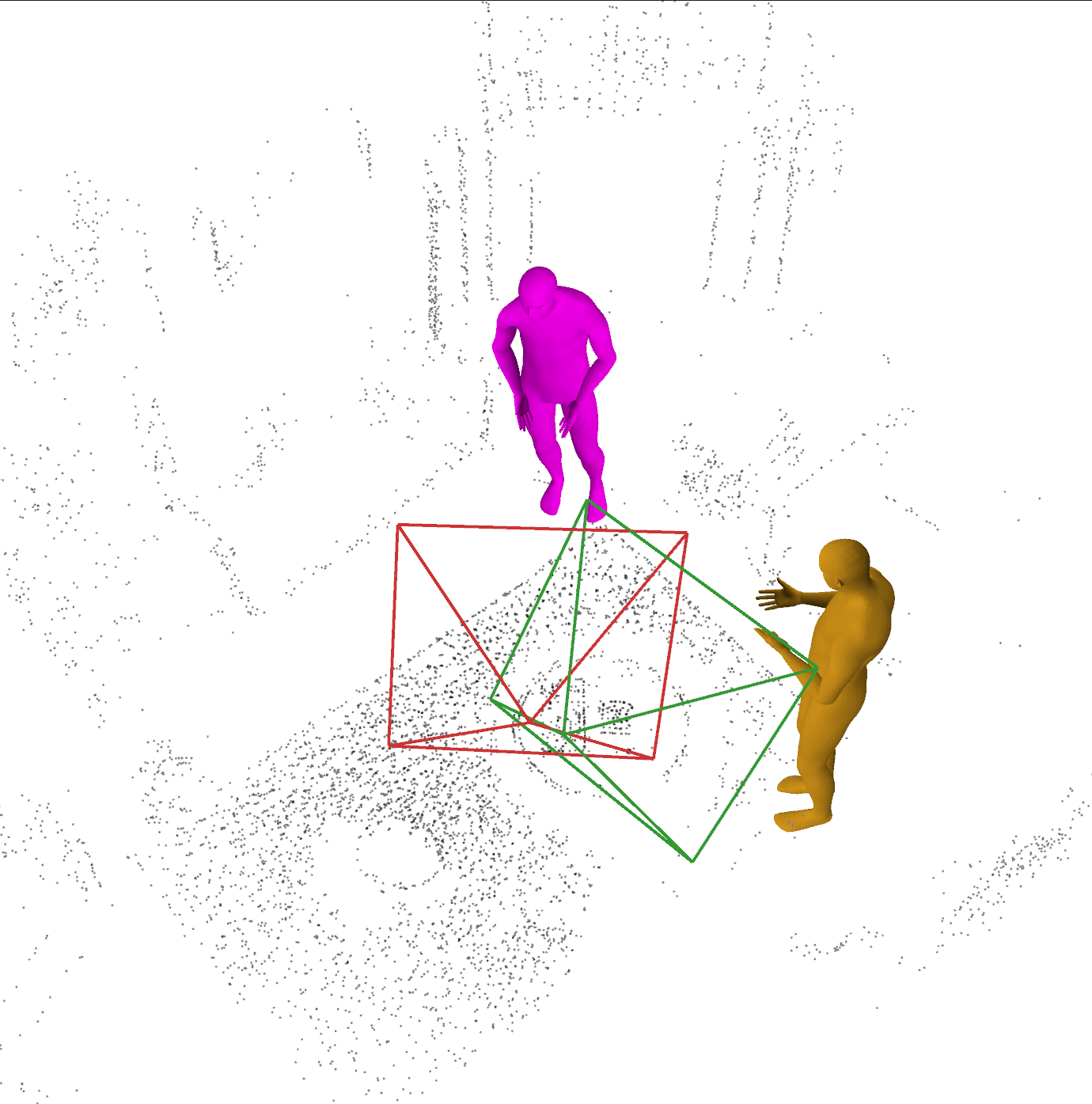}};
    \node(p5) at(p4.east)[anchor=west, xshift=1pt] {\includegraphics[height=\ih]{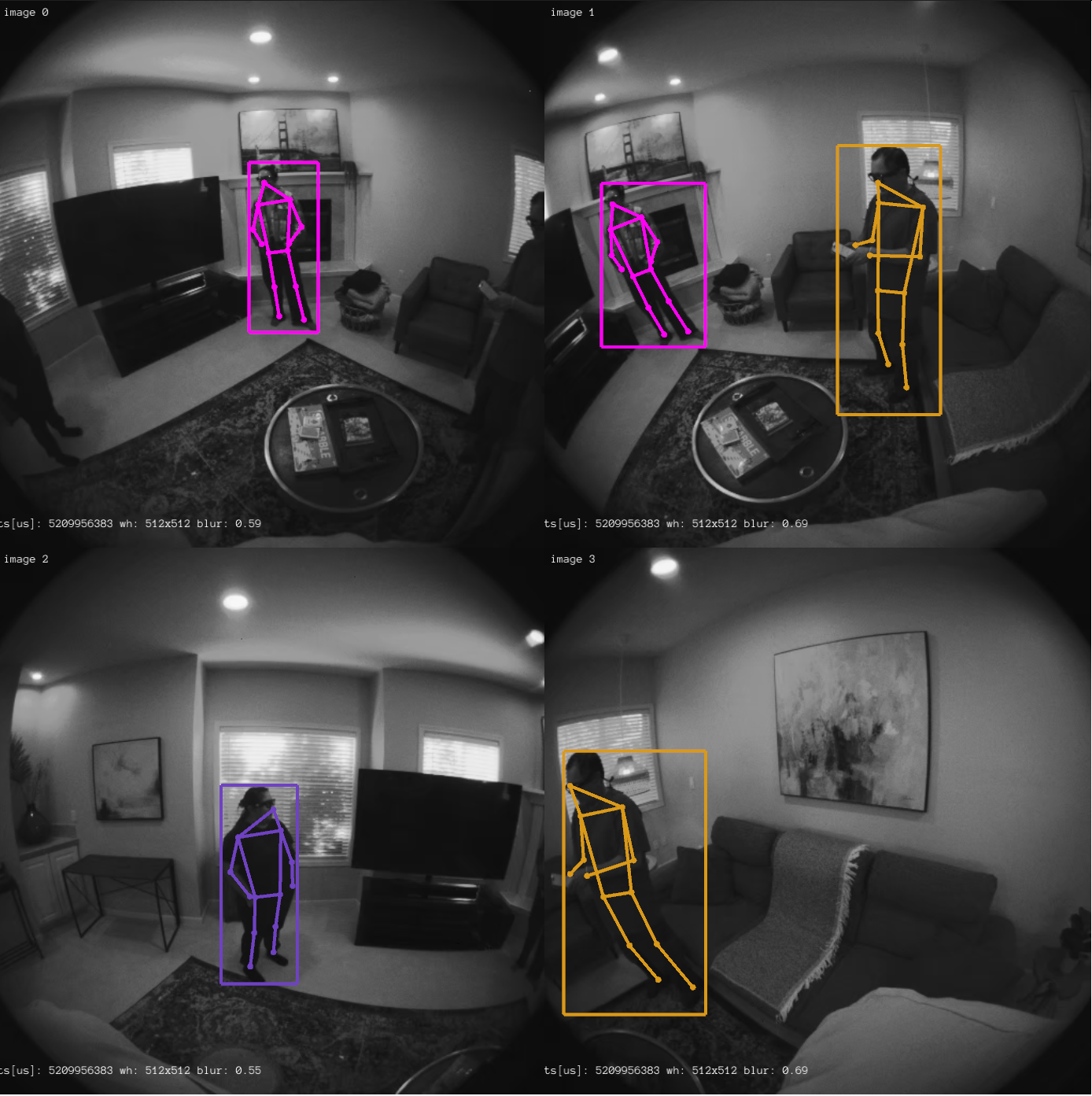}};
    \node(p6) at(p5.east)[anchor=west] {\includegraphics[height=\ih]{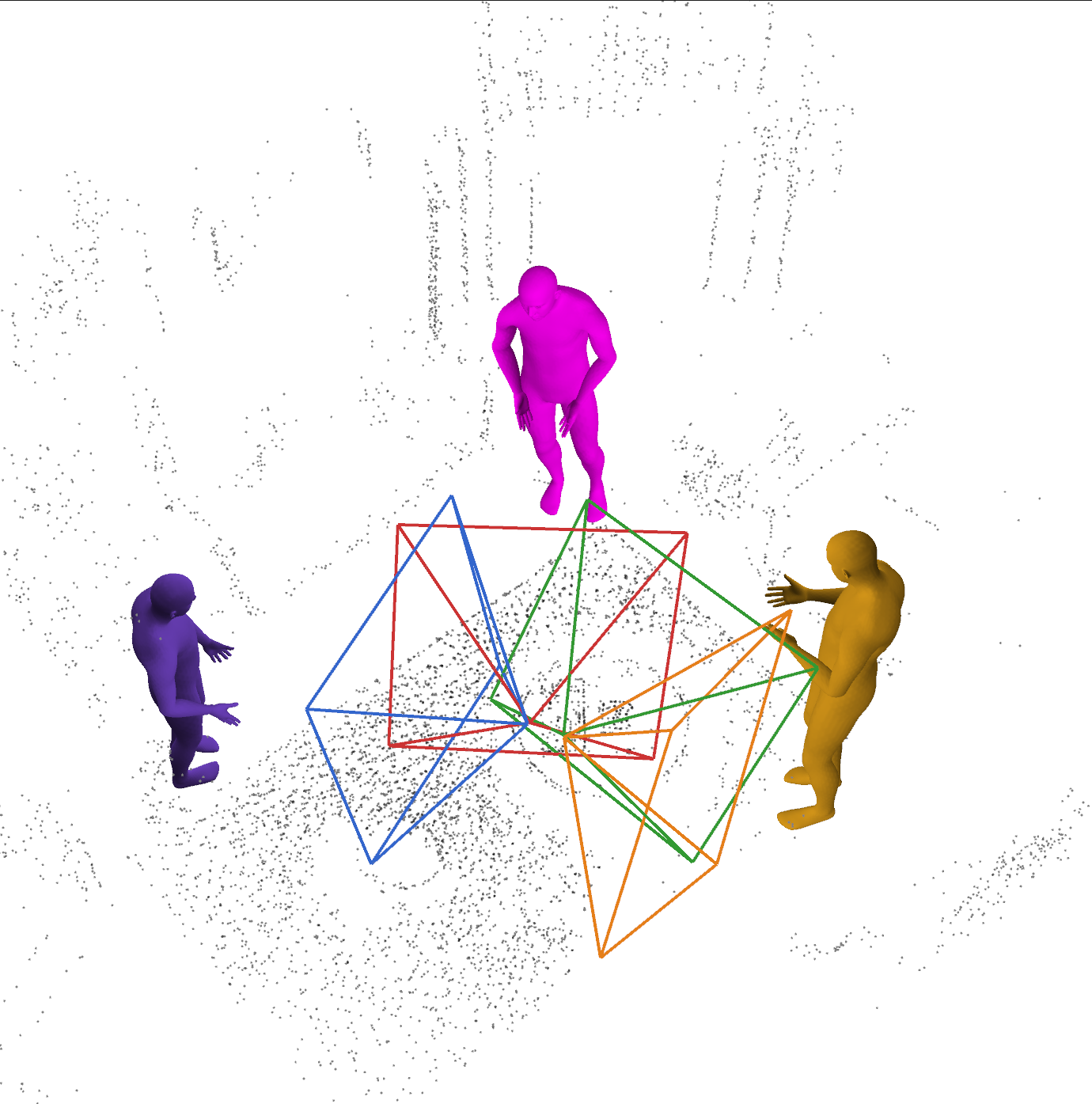}};

\end{tikzpicture}
  \caption{
  \textbf{Tracking coverage versus number of cameras.}
  (\textit{Left}): Distribution of the proportion of people tracked per timestamp against the proportion of time.  Using more cameras clearly shifts mass toward 1.0, meaning all people are tracked more often. In a dynamic social interaction with three other people, average coverage is $47\%$ for \textcolor{blue}{1-cam}, $65\%$ for \textcolor{orange}{2-cam}, and $81\%$ for \textcolor{green}{4-cam}.
\textit{Right}: qualitative examples showing how using more cameras improves the tracking coverage.}
  \label{fig:people_hist}
\end{figure*}

\begin{table*}
\caption{\textbf{Quantitative comparison of \lamp against state-of-the-art algorithms.}
We evaluate \lamp on EMDB~\cite{emdb} and Nymeria~\cite{ma2024nymeria} datasets, where the best results for each dataset is marked in bold, respectively. We refer to GEM~\cite{li2025genmo} for the evaluation results of WHAM~\cite{wham} and GVHMR~\cite{shen2024gvhmr} on EMDB~\cite{emdb} with GT camera poses. The metrics are reported in millimeter (mm) for all MPJPE variants and foot skating (FS), in percentage (\%) for RTE and in $10m/s^3$.}
\label{tab:quantitative}
\centering
\setlength{\tabcolsep}{3pt}
    \newcolumntype{C}[1]{>{\centering\arraybackslash}p{#1}}
    \newcolumntype{L}[1]{>{\raggedright\arraybackslash}p{#1}}
    \newcolumntype{R}[1]{>{\raggedleft\arraybackslash}p{#1}}
    \newcommand{\none}{-}
    \small
    \begin{tabular}{L{30mm} R{16mm} *{4}{R{17mm}} *{3}{R{10mm}}}
    \toprule
    Method                              &Dataset &MPJPE  &PA-MPJPE &$\text{WA-MPJPE}_{100}$ &W-MPJPE &RTE &Jitter &FS \\ \midrule
    WHAM~\cite{wham}                    &EMDB   &81.6  &52.0  &131.1 &\none  &4.1 &21.0 &4.4 \\
    GVHMR~\cite{shen2024gvhmr}          &EMDB   &74.2  &44.5  &109.1 &\none  &1.9 &16.5 &3.5 \\
    GEM~\cite{li2025genmo}            &EMDB   &73.0  &42.5  &69.5 &\none   &0.9 &17.7 &8.6 \\
    PromptHMR~\cite{wang2025prompthmr}  &EMDB   &\textbf{68.1} &\textbf{40.1} &\textbf{63.9} &278.1  &0.4 &16.3 &3.5 \\
    \lamp-mono (ours)                   &EMDB   &82.3  &46.3  &77.8 &\textbf{165.1}    &\textbf{0.2} &\textbf{4.6}  &\textbf{3.2} \\
    \midrule
    PromptHMR~\cite{wang2025prompthmr}  &Nymeria &109.2 &66.0  &101.6 &246.0  &0.11 &114.1 &7.7 \\
    \lamp-mono (ours)                   &Nymeria &92.3  &55.5  &80.4  &203.4  &0.09 &23.8  &\textbf{3.2} \\
    \lamp-mv (ours)                 &Nymeria &\textbf{54.8} & \textbf{37.3} &\textbf{58.7} &\textbf{113.3}  & \textbf{0.05} & \textbf{21.8} & 3.6\\
    \bottomrule
    \end{tabular}
\end{table*}

\subsection{Implementation details}

\paragraph{Model.} \lamp-Net contains three transformer encoder-decoder blocks with $512$ inner dimension.
The input is set to be a 4-second temporal window. With video data at 30~Hz, this amounts to in $W=120$ frames input.
The loss weights are set to be $\lambda_\text{SMPL}=1.0$, $\lambda_\text{3D}=5.0$, $\lambda_\text{V}=1.0$, $\lambda_\text{vel}=20.0$,
The network is trained for 200 epochs on 4 nodes of NVidia H100 GPUs.
Training takes approximately 19 hours to converge.
The inference and evaluation run on a single RTX4090 GPU for real-time inference.

\paragraph{Dataset.}
\lamp-Net is trained with Nymeria dataset~\cite{ma2024nymeria}. The dataset contains 1100 recordings, more than 300 hours of in-the-wild human motion from 264 participants, using multiple Project Aria glasses~\cite{engel2023project}.
The dataset provides ground-truth body motions for subjects using XSens mocap suit, with retargetted SMPL~\cite{smpl} format and we use the observers' Aria glasses to generate training data.
The dataset is then split into 770, 165 and 26 sequences for training, validation and testing. The test set covers one of each 20 activity scenarios to evaluate diverse motion.
For training data preparation, we project the 3D ground-truth body joints onto the observer cameras using the ground-truth intrinsics and extrinsics.
We apply extensive data augmentation to simulate real world noisy 2D detections and occlusions. We also augment the headset motion to further increase the diversity. The data augmentation is further detailed in the supplemental.
In addition, we also evaluate \lamp with EMDB~\cite{emdb}, a widely-adopted dataset in benchmarking 3D human tracking.
Since EMDB is a small dataset with total 40 minutes recordings, we do not train or finetune our model on EMDB.
Therefore, our results on EMDB reflect the zero-shot generalization of \lamp.

\paragraph{Simulate arbitrary multi-camera layout.}
A key benefit of the 3D ray lifting formulation is that \lamp-Net does not require direct pixels or image features for training and inference.
Therefore, we can easily simulate 2D keypoint observations by projecting ground-truth 3D joints into virtual cameras.
This design enables training with arbitrary 3D human motion dataset without video data~\cite{ma2024nymeria,amass}.
Importantly, this training scheme enables large-scale data synthesis~\cite{wham} with diverse observer-camera configurations.
In our experiments, we qualitatively evaluate this feature of \lamp by simulating a large Aria Gen2~\cite{AriaGen2} dataset from the Nymeria dataset which is collected with Aria Gen1~\cite{aria}, and show that the model is able to handle real-world Aria Gen2 dataset, despite never trained with any real Aria Gen2 data. Note that there is a significant update on camera configuration from Aria Gen1 to Aria Gen2.

\paragraph{Metrics.}
We evaluate \lamp using a range of common metrics, mostly derived from the Mean Per-Joint Position Error (MPJPE) error and reported in millimeters (mm).
Here we provide a brief description of different MPJPE variants and propose a new variant W-MPJPE for evaluating motion tracking in metric 3D world.
MPJPE measures the average joint position error after removing any translational misalignment of the root joint (pelvis) w.r.t. ground truth.
Similarly, PA-MPJPE measures the average joint position error after Procrustes alignment with the ground truth.
Both metrics are designed to measure the local similarity of body poses and are computed on a per-frame basis.
$\text{WA-MPJPE}_\text{100}$ on the other hand, is designed to infer the global body pose accuracy.
Due to the common setting with monocular video input, the $\mathbb{SIM}(3)$ alignment is first performed over a 100-frame temporal window before computing the error in order to reduce the impact of scale ambiguity.
For multi-camera headset input, we argue that algorithms should be able to resolve the scale ambiguity and algorithms should be directly evaluated against ground truth without any alignment.
To this end, we introduce a new metric W-MPJPE, which measures the MPJPE without any alignment.
Following previous works~\cite{wang2025prompthmr, wham, slahmr, pace}, we also report jitter, foot skating (FS), and root trajectory error (RTE).

\subsection{Comparison to state-of-the-art}
We conduct evaluations against recent state-of-the-art methods~\cite{wham,shen2024gvhmr,wang2025prompthmr,li2025genmo}. The quantitative comparison on EMDB and Nymeria is shown in \cref{tab:quantitative} and the qualitative comparison on Nymeria is presented in \cref{fig:visual_cmp}.

\begin{figure}[t]
\centering
\begin{subfigure}{0.48\linewidth}
  \includegraphics[width=\linewidth]{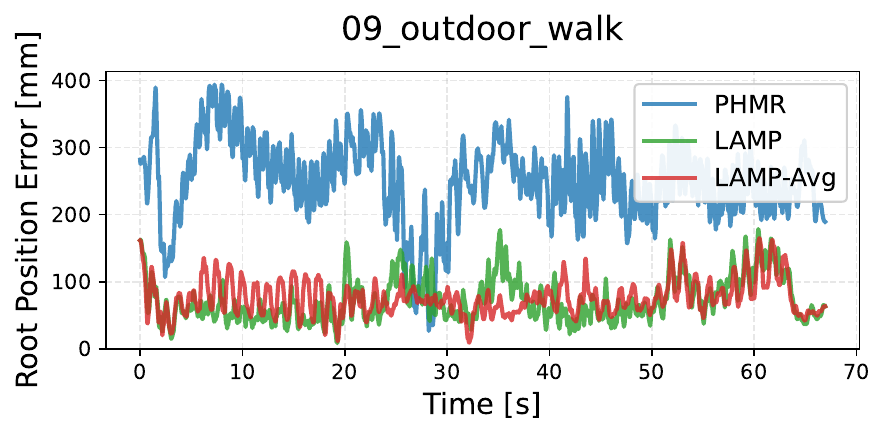}
\end{subfigure}\hfill
\begin{subfigure}{0.48\linewidth}
  \includegraphics[width=\linewidth]{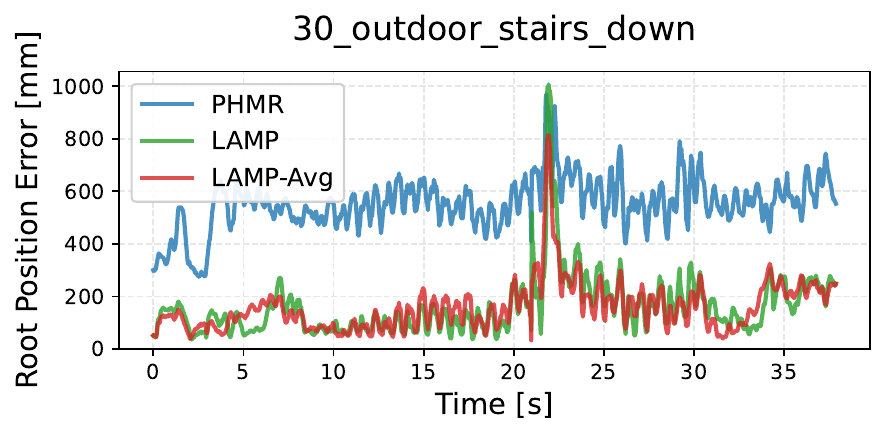}
\end{subfigure}

\begin{subfigure}{0.48\linewidth}
  \includegraphics[width=\linewidth]{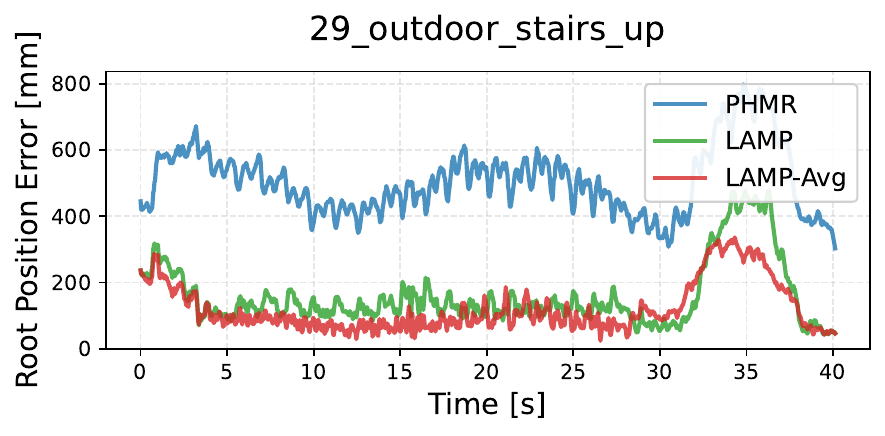}
\end{subfigure}\hfill
\begin{subfigure}{0.48\linewidth}
  \includegraphics[width=\linewidth]{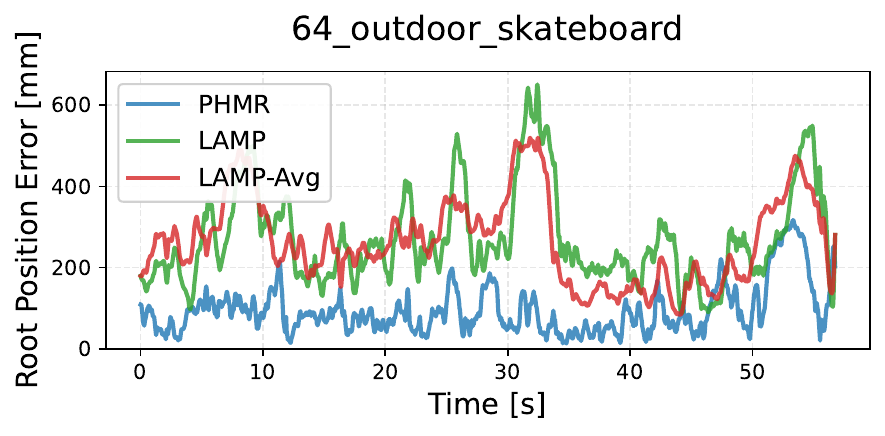}
\end{subfigure}
\caption{\textbf{Root Trajectory Errors on EMDB}. \lamp outperforms PromptHMR on absolute root trajectory accuracy on most sequences. However, on \textit{64\_outdoor\_skateboard}, \lamp shows inferior result which is likely due to the lack of skateboarding activity in the training data. }
\label{fig:emdb_traj_error}
\end{figure}

\paragraph{Evaluation on EMDB.} We first evaluate \lamp on the EMDB-2 test split. EMDB is captured from slowly moving cameras closely following the participant. We use ground-truth camera extrinsics for all the methods and run \lamp \textit{with single-view input} for direct comparison to other methods. Overall, \METHOD achieves highest RTE, W-MPJPE and Jitter by a significant margin, indicating superior world-space localization accuracy. \Cref{fig:emdb_traj_error} also shows the trajectory error overtime comparing \lamp and PromptHMR. In turn, \METHOD falls notably behind on MPJPE, PA-MPJPE and $\text{WA-MPJPE}_\text{100}$.  We believe this can be attributed to our design choice of collapsing raw images into 3D rays to facilitate multi-view aggregation and continuous tracking over time, capabilities which are not captured by these metrics and on this dataset. Please refer to our supplementary materials for the qualitative results on EMDB.

\begin{table*}
\caption{\textbf{Ablation study.}
The design choices (first four columns) mean the following.
By \textit{posed}, we transform 3D rays by camera poses;
by \textit{smooth}, per-frame estimation is averaged over a sliding window;
by \textit{simulated}, we simulate 2D keypoint detections from ground truth;
and by \textit{multiview}, we use four-camera~\cite{AriaGen2} observations instead of monocular video.
Experiments is done with the Nymeria~\cite{ma2024nymeria} datasets.
All MPJPE variants and foot skating (FS) are in millimeter (mm), RTE in percentage (\%) and jitter in $10m/s^3$.}
\label{tab:ablation}
\centering
\setlength{\tabcolsep}{3pt}
    \newcolumntype{C}[1]{>{\centering\arraybackslash}p{#1}}
    \newcolumntype{L}[1]{>{\raggedright\arraybackslash}p{#1}}
    \newcolumntype{R}[1]{>{\raggedleft\arraybackslash}p{#1}}
    \newcommand{\none}{-}
    \small
    \begin{tabular}{L{8mm} *{2}{C{10mm}} *{2}{C{13mm}} | C{14mm} *{3}{C{18mm}} *{3}{C{8mm}} }
    \toprule
    variant &posed &smooth  &simulate  &multiview  &MPJPE  &PA-MPJPE  &$\text{WA-MPJPE}_{100}$ &W-MPJPE &RTE &Jitter &FS \\ \midrule
    \textit{$var_0$} &     &       &          &    & 98.5 & 57.5 & 95.8 & 296.3 & 0.50 & 93.1 & 6.1 \\
    \textit{$var_1$} &\checkmark &  &  &           & 98.3 & 57.6 & 89.3 & 209.6 & 0.09 & 91.7 & 5.5 \\
    \textit{$var_2$} &\checkmark &\checkmark &  &  & 92.3 & 55.5 & 80.4 & 203.4 & 0.09 & 23.8 & 3.2 \\
    \textit{$var_3$} &\checkmark &\checkmark &\checkmark  & & 60.4 & 39.2 & 71.7 & 199.8 & 0.08 & 21.4 & 3.5 \\
    \textit{$var_4$} &\checkmark &\checkmark &  &\checkmark & 54.8 & 37.3 & 58.7 & 113.3 & 0.05 & 21.8 & 3.6\\
    \textit{$var_5$} &\checkmark &\checkmark &\checkmark &\checkmark & 52.0 & 34.8 & 58.2 & 111.5 & 0.05 & 21.4 & 3.5\\
    \bottomrule
    \end{tabular}
\end{table*}

\vspace{-5mm}
\paragraph{Evaluation on Nymeria.} We compare \METHOD with PromptHMR~\cite{wang2025prompthmr} on the Nymeria dataset. Nymeria differs from EMDB in 2 key aspects: 1) it contains longer sequences (15 mins versus 1 min per sequence) allowing better evaluation over long-term accuracy; 2) the observer camera contains natural egocentric head motions with frequent rotations. For comparison we adapt PromptHMR to take ground-truth camera poses, rectify, and process the sequences in 1200 frame-chunks. We observe that \METHOD notably outperforms PromptHMR on all metrics even in the monocular configuration, where multiview input gives additional boost to accuracy as shown in~\cref{fig:visual_cmp}.
Note that for the multi-camera setup we use the keypoint detections from 1 RGB and 2 SLAM cameras from Project Aria glasses (Gen1). This is further discussed in Sec.~\ref{ssec:ablations}. More qualitative results are presented in the supplementary.

\paragraph{Qualitative evaluation on Aria Gen2.}
To show that our training method enables generalizaiton to different camera layouts, we recorded test sequences with the Project Aria Gen2 headset~\cite{AriaGen2}, which mounts 4 CV cameras: a forward-facing stereo pair with substantial overlap plus two side-view cameras, yielding a $\sim\!270^\circ$ field of view. \lamp exploits the known intrinsics/extrinsics to fuse all views; as illustrated in \cref{fig:teaser}, it persistently tracks the people over long trajectories (\textit{Left}) and tracks multiple people in real-time across the wide FoV (\textit{Right}). To quantify the effect of camera count on coverage, we analyze four 5-minute sequences from the Aria Gen2 Pilot dataset~\cite{kong2025aria}, comparing 1-, 2-, and 4-camera settings. \Cref{fig:people_hist} shows that adding views substantially increases per-timestamp coverage in dynamic social interactions, confirming the benefit of the multi-camera rig.

\subsection{Ablation studies}
\label{ssec:ablations}
We conduct ablation studies on the Nymeria dataset. We pre-process the dataset to evaluate only frames with sufficient visibility of the tracked person -- we observe that the dataset contains complex activities and interactions, and thus -- in contrast to EMDB -- the tracked person is not always visible and can be temporarily occluded by e.g. walls or furniture. We achieve this by running an off-the-shelf 2D bounding box detector~\cite{wu2019detectron2} for people on the full video, and filter out any frames in which no person is detected with a 2D bounding-box IoU of at least 0.4 compared to surrounding boxes of the projected 3D joints.

\Cref{tab:ablation} reports the effect of each design choice for LAMP. The first row (\textit{$var_0$}) is the result of LAMP when \textit{not using known camera motion}, and instead lifting rays into the moving observer frame of reference. Adding LAMP's 3D Ray-Lifting (\textit{$var_1$}) significantly improves all Global Metrics, as well as RTE and foot-skating. This is expected, and indeed a core benefit of our method which allows to fully leverage known observer-motion for accurate 3D localization of the observed person. Averaging overlapping windows further improves all metrics (\textit{$var_2$}).
To evaluate the impact of multiple cameras, we use Aria Gen1 multi-view camera rig with 2D keypoints detected by ViTPose~\cite{vitpose} (\textit{$var_4$}). In addtion, we include results with simulated 2D keypoints for both monocular and multi-view experiments (\textit{$var_3$} and \textit{$var_5$}). The comparison between \textit{$var_2$} and \textit{$var_4$} shows the improvement by using all 3 cameras vs. when using only a monocular view. Again, using multi-view input yields significant improvement across all metrics except jitter and foot-sliding. An additional observation is that the sim-to-real delta of multi-view inputs (i.e., between \textit{$var_4$} and \textit{$var_5$} is significantly smaller than that of using monocular inputs (i.e., between \textit{$var_2$} and \textit{$var_3$}), which suggests, although LAMP-Net is trained with simulation only, using multi-view inputs together with our extensive data augmentation closes the sim-to-real gap notably. Please refer to our supplementary materials for additional experiments on camera pose and 2D keypoints sensitivity, runtime, and tracking performance.

\section{Conclusion}
\label{sec:conclusion}
\METHOD introduces a new approach to tracking people and their body-motion in 3D metric world space. By accepting known 6DoF motion and calibration as readily-available input modality, \METHOD provides a simple and flexible solution to disentangling observer- and target-motion, as well as to fuse observations from multi-view video input. By further abstracting raw pixels into 2D rays early-on, \METHOD can easily be re-trained for arbitrary rig-layouts facilitating cross-device usage. We demonstrate \METHOD's superior performance in the targeted scenario of in-the-wild egocentric observations of surrounding people -- as supposed to intentional/staged capture from static or hand-held cameras.

\textbf{Limitations} \METHOD depends on accurate and reliable 6~DoF tracking as input modality, and requires multi-view input to unfold its full potential. While this is commonly available for modern egocentric devices, it does prevent \METHOD from being used effectively for monocular mobile phone captures or common internet-sourced recordings. Besides, including more comprehensive pixel-derived information will potentially improve the local human pose accuracy.

\clearpage
\section*{Acknowledgements} We would like to thank Federica Bogo, Bharat Bhatnagar, Jinlong Yang, Yuanlu Xu, David Cruso, and Daniel DeTone for their valuable support and fruitful discussions throughout the course of this project.
\bibliographystyle{ieeenat_fullname}
\bibliography{main}

\clearpage

\twocolumn[{
  \renewcommand\twocolumn[1][]{#1}%
  \maketitlesupplementary
  \vspace{-2mm}
  \begin{center}
    \captionsetup{type=figure}
    \begin{subfigure}[b]{0.32\textwidth}
      \centering
      \includegraphics[width=\linewidth]{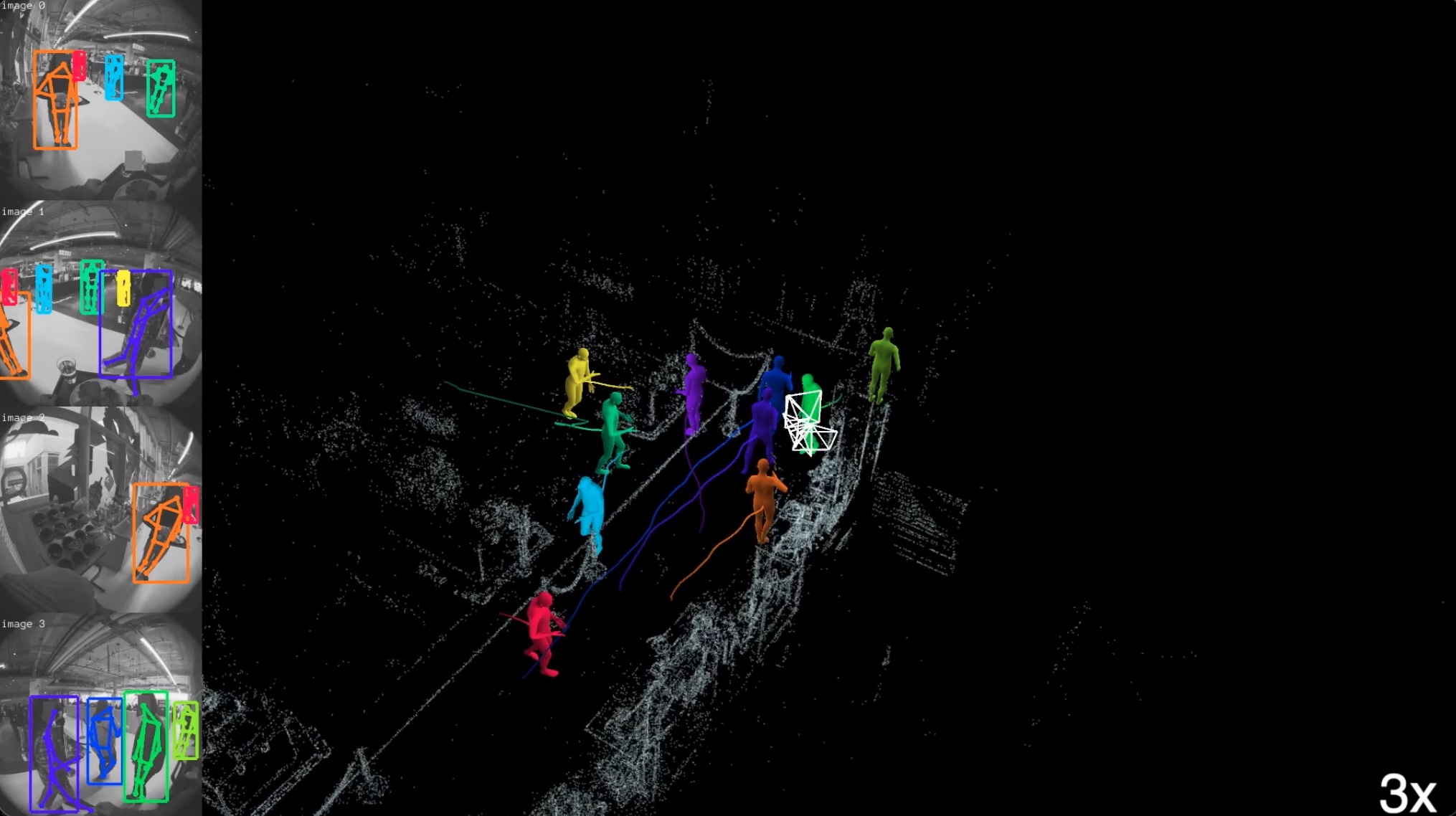}
      \caption{Scene 1 - Cafeteria.}
    \end{subfigure}
    \hfill
    \begin{subfigure}[b]{0.32\textwidth}
      \centering
      \includegraphics[width=\linewidth]{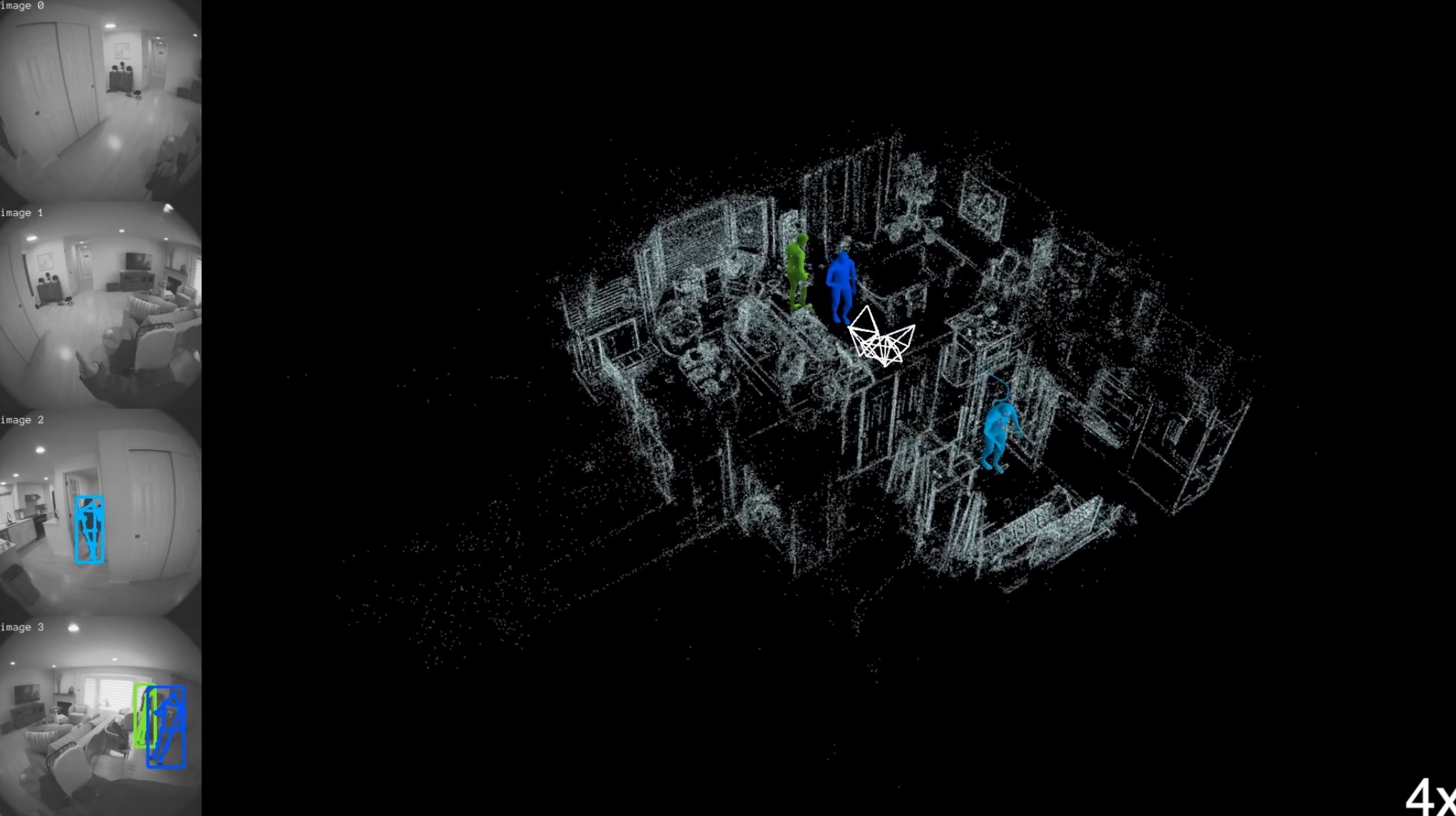}
      \caption{Scene 2 - Apartment.}
    \end{subfigure}
    \hfill
    \begin{subfigure}[b]{0.32\textwidth}
      \centering
      \includegraphics[width=\linewidth]{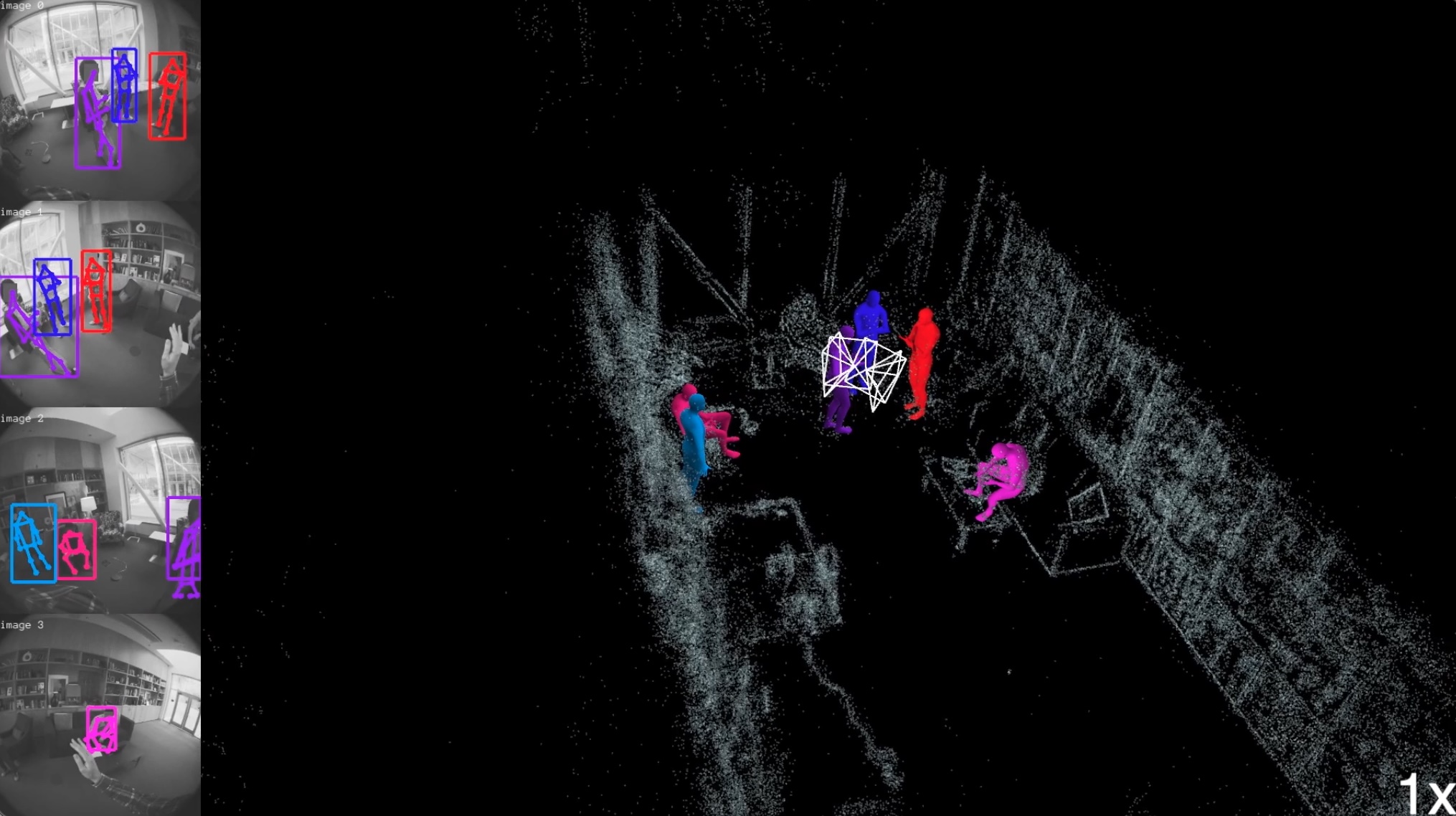}
      \caption{Scene 3 - Library.}
    \end{subfigure}

    \caption{\textbf{Real-time real-world demo with Aria Gen 2~\cite{AriaGen2} headset}. We show 3 scenarios with Aria Gen 2 headset to showcase \lamp in tracking multiple people for casual social activities. Note the algorithm is trained on simulation and tested with real-world data.}
    \label{fig:suppl_demo_video}
  \end{center}
  \vspace{2mm}
}]

\section{Supplementary Video}
\label{sec:supp_qualitative}
In order to provide better visual assessment of \lamp in both 3D grounding and temporal consistency,
we provide videos to visualize the algorithm outputs.
We kindly refer the readers to the supplementary video.
Following we briefly describe the video content.

\subsection{Evaluation on public datasets}
\begin{figure*}
    \centering
    \includegraphics[width=0.98\linewidth]{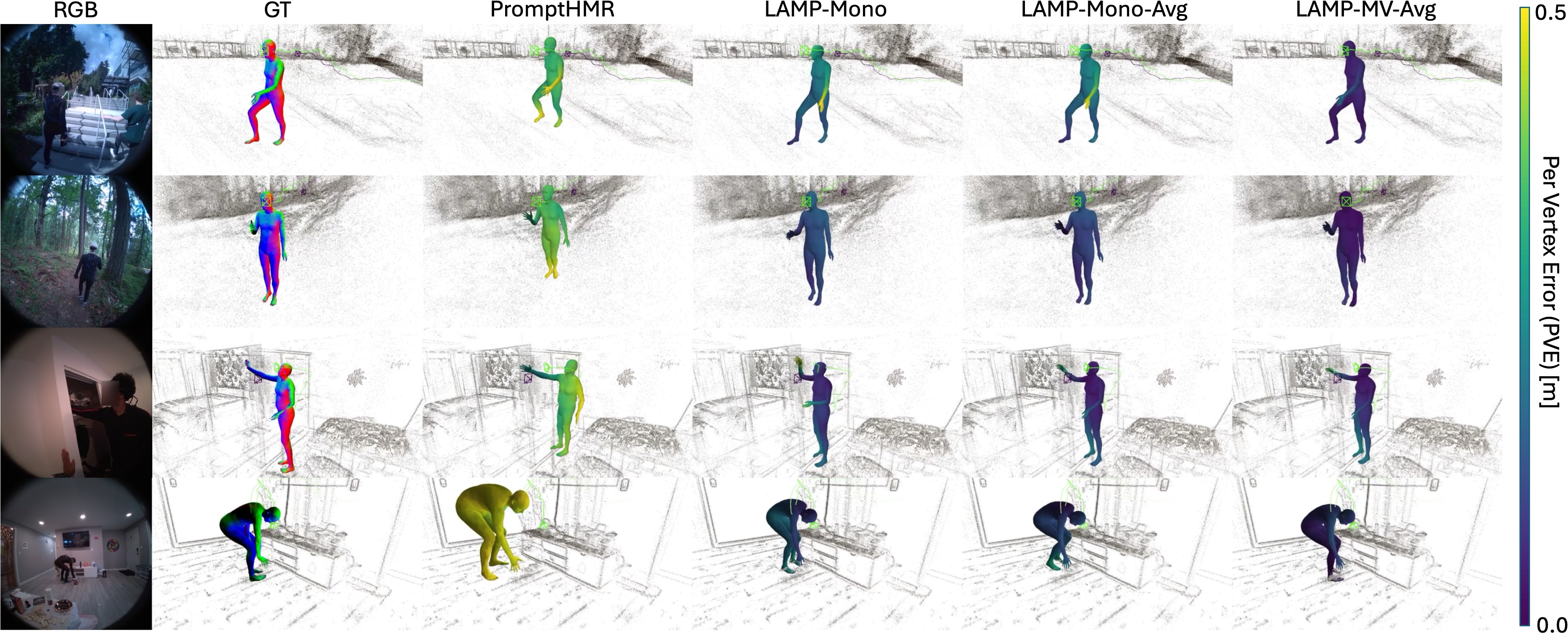}
    \caption{\textbf{Qualitative comparisons on Nymeria}. We compare PromptHMR~\cite{wang2025prompthmr} with LAMP variants, and show the benefits of using the temporal averaging and multi-view inputs. The vertices are colored by Per Vertex Error (PVE) in the world coordinate. Please refer to the supplementary video to view the full comparison.}
    \label{fig:supp_nymeria}
\end{figure*}

\begin{figure*}
    \centering
    \includegraphics[width=0.96\linewidth]{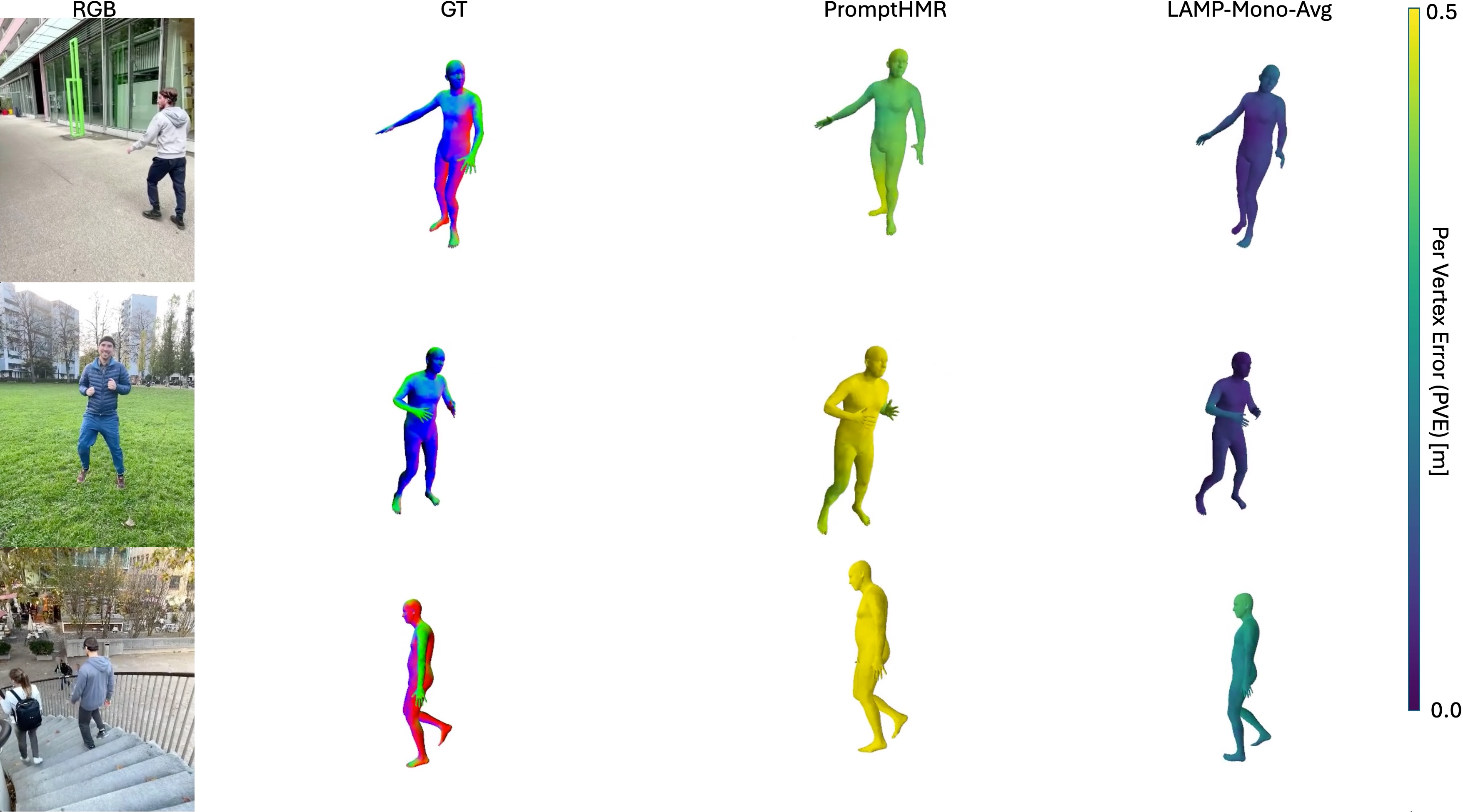}
    \caption{\textbf{Qualitative comparison on EMDB}. We compare LAMP-Mono-Avg with PromptHMR~\cite{wang2025prompthmr} using the monocular video input from EMDB. Note the result from \lamp shows the zero-shot generalization without training on EMDB. Please refer to the supplementary video for full assessment.}
    \label{fig:supp_emdb}
\end{figure*}

We show further comparison of \lamp with the state-of-the-art method PromptHMR~\cite{wang2025prompthmr}.
\Cref{fig:supp_nymeria} shows the screenshots for the result on the Nymeria dataset~\cite{ma2024nymeria}
and \cref{fig:supp_emdb} shows the screenshot for the result on the EMDB dataset~\cite{emdb}.
The video rendering uses the same color coding as used in \cref{fig:visual_cmp},
which corresponds to the Per Vertex Error (PVE) computed in the world coordinate.
To provide 3D visual reference, we show the trajectory of observing camera in purple.
For Nymeria dataset, the target person also wears a headset, which is localized in the same metric coordinates as the observing cameras.
We therefore show their headset trajectories in green to highlight the accuracy of 3D body grounding.
If the tracking is accurate, the position of estimated head should match the green headset.
In the video, we show that while \lamp performs on par with PromptHMR in estimating the local body poses,
\lamp also consistently outperforms PromptHMR in grounding the body motion in metric 3D world.
This proves the effectiveness of \lamp in factoring out the headset motion in its formulation.
Note that all results from \lamp on EMDB are zero-shot, showing the effectiveness of using multi-view temporal posed 3D rays directly in \lamp-Net training.


\subsection{Real-time real-world demo}
In addition to evaluation on public benchmark, we show live demos running in real-time with real world data.
To this end, we use the Project Aria Gen 2~\cite{AriaGen2} headset and collect 3 diverse scenarios where multiple people
perform casual activities at home, in the office and at the cafeteria (c.f., \cref{fig:suppl_demo_video} for screenshots of each demo).
For real-time demo, we use a lightweight MHR model~\cite{MHR2025} instead of SMPL~\cite{smpl}.
This change only requires minor alter to our algorithm to output MHR parameters.
Note \lamp is only trained with simulated Aria Gen 2 data for real-world testing, which is benefit from the ray-based formulation.
The demos highlight real-world challenges, with rapid motion, natural occlusions and 2D observations of the same people constant switching across different cameras over time. Leveraging the spatio-temporal posed 3D ray fusion paradigm, \lamp is able to handle these challenges well.

\section{Additional Experiments}
\noindent\textbf{Camera pose sensitivity}
In the paper we compare \textit{monocular} LAMP against baselines on EMDB using \textit{GT camera poses}. Here we further perturb GT poses with temporally correlated SE(3) noise (sampled every 10\,s interval and smoothly ramped within the interval), sweeping translation/rotation from $2$--$8$\,cm and $0.02^\circ$--$0.2^\circ$ (Tab.~\ref{tab:cam_noise}), which are in the range of SOTA academic VIO systems like OKVIS2~\cite{leutenegger2022okvis2}.
We also include the results using poses from DROID-SLAM~\cite{droid}.
As expected, LAMP degrades with noisy poses, but remains superior to PromptHMR (PHMR) even when PHMR uses GT poses.
The advantage remains when both methods use DROID-SLAM poses.
The results suggest LAMP yields both a higher upper bound with accurate poses and robustness under realistic pose errors.
It is important to point out that industrial VIO/SLAM, such as Aria localization algorithm used in our work, are substantially more accurate than academia solutions as reported in LaMAria benchmark~\cite{lamaria} and yield order of magnitude lower noise than the value used in Tab.~\ref{tab:cam_noise}. This motivates our design to disentangle camera motion and human motion.

\begin{table}[t]
\centering
\begin{minipage}{0.56\columnwidth}
  \centering
  \centering
\footnotesize
\setlength{\tabcolsep}{1pt}
\renewcommand{\arraystretch}{1.0}
\begin{tabular}{lccc}
\toprule
\textbf{Method} & \textbf{Cam. Pose} & \textbf{W-MPJPE} & \textbf{Jitter} \\
\midrule
PHMR & GT & 278.1  & 16.3 \\
LAMP & GT & 165.1  & 4.6 \\
LAMP & 2cm/0.02$^\circ$ & 165.7  & 4.7 \\
LAMP & 4cm/0.05$^\circ$ & 170.2  & 4.8 \\
LAMP & 6cm/0.1$^\circ$ & 181.9  & 4.8 \\
LAMP & 8cm/0.2$^\circ$ & 212.3  & 5.0 \\
\midrule
PHMR & DROID & 294.3  & 17.1 \\
LAMP & DROID & 273.9  & 5.2 \\
\bottomrule
\end{tabular}

  \vspace{-6pt}
  \captionof{table}{\textbf{Ablation on camera poses.}}
  \label{tab:cam_noise}
\end{minipage}\hfill
\begin{minipage}{0.40\columnwidth}
  \centering
  \centering
\footnotesize
\setlength{\tabcolsep}{1pt}
\renewcommand{\arraystretch}{1.0}
\begin{tabular}{ccc}
\toprule
\textbf{2D Kp} & \textbf{W-MPJPE} & \textbf{Jitter} \\
\midrule
ViTPose-S & 170.7  & 4.7 \\
ViTPose-B & 168.2  & 4.6 \\
ViTPose-L & 165.4  & 4.6 \\
ViTPose-H & 165.1  & 4.6 \\
\midrule
$\sigma = 1px$ & 165.1 & 4.6 \\
$\sigma = 5px$ &  165.9 & 4.6 \\
$\sigma = 10px$ & 166.8 & 4.8 \\
\bottomrule
\end{tabular}

  \vspace{-6pt}
  \captionof{table}{\textbf{Ablation on 2D KPs.}}
  \label{tab:kp_noise}
\end{minipage}
\vspace{-6pt}
\end{table}

\noindent\textbf{2D keypoint sensitivity}
We ablate different ViTPose backbones (S--H) and added Gaussian noise to ViTPose-H on EMDB in Tab.~\ref{tab:kp_noise}.
The results show that LAMP degrades minimally under significant noise, confirming tolerance to both limited model capacity and pixel-level jitter.
The robustness benefits from the extensive data augmentation during training as described in submission.

\noindent\textbf{Runtime, latency clarification}
LAMP comprises 3 components: YoloX-S 2D detection, ViTPose-S 2D keypoints and LAMP-Net. Fig.~11 breaks down the runtime on RTX~4090 against number of tracklets.
With 10 tracklets, LAMP runs at $\sim$12.5\,Hz, whereas PHMR runs at $\sim$6\,Hz with only 1 tracklet. Fig.~12 shows the tradeoff of delay over accuracy/stability with temporal smoothing, allowing applications to choose latency budget.

\noindent\textbf{Tracking Performance}
We clarify that ``tracking'' in LAMP emphasizes \emph{world-grounded motion estimation} rather than long-term re-identification. Standard MOT metrics are unsuitable here as our benchmarks provide ground truth for only a single subject. To address tracking concerns, we computed \textit{3D tracking recall} on Nymeria (0.25m threshold at the pelvis, ignoring IDs); LAMP achieves 90.3\%, confirming high coverage. Additionally, Fig.~6 quantifies multi-camera benefits via tracking coverage analysis. We further illustrate high recall and stable identity association under rapid camera motion and partial view dropouts in the supplementary video (e.g., crowded cafeteria).

\begin{figure}
\centering
\begin{minipage}{.5\columnwidth}
  \centering
  \includegraphics[width=.9\linewidth]{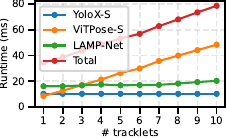}
  \vspace{-6pt}
  \caption*{\scriptsize Fig. 11 \textbf{LAMP System Runtime}}
\end{minipage}%
\begin{minipage}{.5\columnwidth}
  \centering
  \includegraphics[width=.9\linewidth]{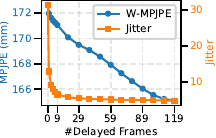}
  \vspace{-6pt}
  \caption*{\scriptsize Fig. 12 \textbf{Temporal Averaging}}
\end{minipage}
\vspace{-6pt}
\end{figure}

\section{Data Augmentations}
\label{sec:supp_data_aug}
We conduct extensive data augmentations to improve the robustness of \lamp-Net on the real-world 2D keypoints with two families of augmentations: temporally correlated noise on visible joints and structured masking that removes observations in realistic patterns.
\paragraph{Noise} We add temporally correlated Gaussian jitter per joint/keypoint track to model detector behavior. The correlation is high so noise evolves smoothly over time. We also add per-frame noise of which the noise magnitude is smaller. In addition, we also make distal joints $1.5\times$ noisier (e.g., wrists/ankles) for which the detections are usually less stable.
\paragraph{Masking} We compose two simple masks to simulate occlusions and view dropouts. First, we random sample time spans with 10 to 20 frames per span. For each span we sample the number of active views, biasing toward multi‑view frames. To cover the monocular case, with a small probability, we force the entire input snippet to have exactly one active view. Second, Within each active view we mask joints in contiguous temporal bursts to simulate self‑occlusion and tracking drops. Bursts last 10 to 20 frames with 1 to 4 joints. We directly set both the coordinates and confidence to $0$ when the points are masked out. In addition, we use a higher probability to mask out feet to simulate real world ego-centric scenarios.

To retain peak accuracy on clean data while gaining robustness, we mix two clip types in training, i.e., clips without noise or masking and clips with light jitter and no masking.

\end{document}